\documentclass[10pt,twocolumn,letterpaper]{article}
\usepackage{iccv}
\usepackage{times}
\usepackage{epsfig}
\usepackage{graphicx}
\usepackage{amsmath}
\usepackage{amssymb}

\usepackage[utf8]{inputenc}
\usepackage[T1]{fontenc}   
\usepackage{url}          
\usepackage{booktabs}      
\usepackage{amsfonts}      
\usepackage{nicefrac}    
\usepackage{microtype}     

\usepackage{bbding}
\usepackage{pifont}

\usepackage{bm}
\usepackage{comment}
\usepackage{enumerate} 
\usepackage{enumitem}
\usepackage{mathrsfs}
\usepackage{makeidx}      
\usepackage{graphicx}       
\usepackage{multicol}       
\usepackage{subfigure}
\usepackage{epsfig,latexsym}
\usepackage{psfrag}
\usepackage[ruled,vlined]{algorithm2e}
\usepackage{algorithmic}
\usepackage{makecell}
\usepackage[table]{xcolor}
\usepackage{fancyhdr}
\usepackage{multirow}
\usepackage{rotating}
\usepackage{color}

\definecolor{lightgray}{rgb}{.93,.93,.93}
\definecolor{deepred}{rgb}{0.698,0.133,0.133}
\definecolor{blue}{rgb}{0.0,0.0,1.0}

\newenvironment{myitemize}{\begin{list}{$\bullet$}
{\setlength{\topsep}{1mm}
	\setlength{\itemsep}{0.1mm}
	\setlength{\parsep}{0.0mm}
	\setlength{\itemindent}{0mm}
	\setlength{\partopsep}{0mm}
	\setlength{\labelwidth}{1mm}
	\setlength{\leftmargin}{4mm}}}{\end{list}}

\newcommand{\mf}{\mathbf}

\newcommand{\mr}{\mathrm}

\usepackage[breaklinks=true,bookmarks=false,colorlinks,urlcolor=red,citecolor=blue,linkcolor=blue]{hyperref}

\iccvfinalcopy 

\def\iccvPaperID{****} 

\ificcvfinal\fi

\begin{document}

\title{Heterogeneous Forgetting Compensation for Class-Incremental Learning}

\author{Jiahua Dong\textsuperscript{1, 2, 3}\footnotemark[1]~, Wenqi Liang\textsuperscript{1, 2, 3}\footnotemark[1]~, Yang Cong\textsuperscript{4}\footnotemark[2]~, Gan Sun\textsuperscript{1, 2} \\
\textsuperscript{1}State Key Laboratory of Robotics, Shenyang Institute of Automation, \\ Chinese Academy of Sciences, Shenyang, 110016, China.\footnotemark[3]\\
\textsuperscript{2}Institutes for Robotics and Intelligent Manufacturing, \\ Chinese Academy of Sciences, Shenyang, 110169, China. \\
\textsuperscript{3}University of Chinese Academy of Sciences, Beijing, 100049, China. \\
\textsuperscript{4}South China University of Technology, Guangzhou, 510640, China. \\
{\tt\small \{dongjiahua1995,\;liangwenqi0123,\;congyang81,\;sungan1412\}@gmail.com}
}

\maketitle

\renewcommand{\thefootnote}{\fnsymbol{footnote}}
\footnotetext[1]{Equal contributions. 
\footnotemark[2]The corresponding author is Prof. Yang Cong.
$~~$\indent\footnotemark[3]This work was supported in part by the National Nature Science Foundation of China under Grant 62127807, 62273333 and 62133005.}

\ificcvfinal\thispagestyle{empty}\fi

\begin{abstract}
Class-incremental learning (CIL) has achieved remarkable successes in learning new classes consecutively while overcoming catastrophic forgetting on old categories. However, most existing CIL methods unreasonably assume that all old categories have the same forgetting pace, and neglect negative influence of forgetting heterogeneity among different old classes on forgetting compensation. To surmount the above challenges, we develop a novel \underline{\textbf{H}}eterogeneous \underline{\textbf{F}}orgetting \underline{\textbf{C}}ompensation (\textbf{HFC}) model, which can resolve heterogeneous forgetting of easy-to-forget and hard-to-forget old categories from both representation and gradient aspects. Specifically, we design a task-semantic aggregation block to alleviate heterogeneous forgetting from representation aspect. It aggregates local category information within each task to learn task-shared global representations. Moreover, we develop two novel plug-and-play losses: a gradient-balanced forgetting compensation loss and a gradient-balanced relation distillation loss to alleviate forgetting from gradient aspect. They consider gradient-balanced compensation to rectify forgetting heterogeneity of old categories and heterogeneous relation consistency. Experiments on several representative datasets illustrate effectiveness of our HFC model. 
The code is available at \url{https://github.com/JiahuaDong/HFC}. 
\end{abstract}

\section{Introduction}
Class-incremental learning (CIL) \cite{Rebuffi_2017_CVPR, Dong_2022_CVPR, 9891836} has attracted appealing attentions recently by accumulating previous learned experience to learn new classes incrementally. It plays an indispensable role in developing a large number of intelligent learning systems, such as autonomous driving \cite{8569992} and automated surveillance \cite{6482546}. When learning new classes continuously under the settings of limited memory to replay all previous data of old classes \cite{Tiwari_2022_CVPR}, these CIL systems heavily suffer from forgetting on old classes caused by severe class imbalance between old and new categories \cite{Rebuffi_2017_CVPR, joseph2022Energy, DongXin_2022_CVPR}. To surmount catastrophic forgetting, a growing amount of CIL methods mainly perform knowledge distillation \cite{44873_Distilling} to preserve past experience \cite{8107520, Dong_2022_CVPR, liu2022deja}; introduce a rehearsal strategy to replay part of old data \cite{Rebuffi_2017_CVPR, 10.1007/978-3-030-58565-5_6}; or dynamically expand network architectures \cite{10.5555/3454287.3455512, pmlr-v80-serra18a, Yan_2021_CVPR}.

\begin{figure}[t]
\centering
\includegraphics[trim = 75mm 54mm 75mm 55mm, clip, width=215pt, height=145pt]
{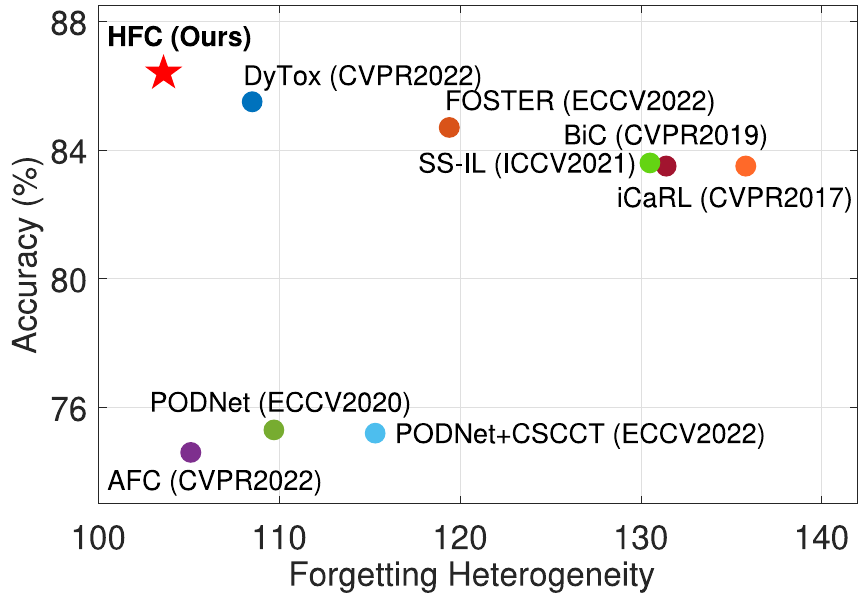}
\caption{Forgetting heterogeneity versus accuracy on CIFAR-100 \cite{krizhevsky2009learning} when the backbone is ViT-Base and number of tasks is $5$. } 
\label{fig: motivation}
\vspace{-12pt}
\end{figure}

However, most existing CIL methods \cite{Rebuffi_2017_CVPR, Tiwari_2022_CVPR, Yan_2021_CVPR, Douillard_2022_CVPR} neglect heterogeneous forgetting speeds of different old classes. They unrealistically assume that all old classes suffer from the same degree of catastrophic forgetting, and compensate forgetting for each old classes equally and independently. Such impracticable assumption enforces existing CIL models \cite{Rebuffi_2017_CVPR, 10.5555/3305890.3306093, zhao2022rbc} to suffer from imbalanced gradient optimization among different old classes, thus favoring more forgetting compensation for hard-to-forget old classes while neglecting those easy-to-forget old classes (\emph{i.e.}, heterogeneous forgetting). More importantly, such forgetting heterogeneity can significantly worsen the forgetting on hard-to-forget old categories, when new streaming classes becomes part of old categories continually. For example, some old classes with various modalities and appearances (\emph{e.g.}, \texttt{person} and \texttt{car}) in autonomous driving \cite{8569992} are more difficult to explore task-shared representations across different incremental tasks, when compared with other hard-to-forget old classes (\emph{e.g.}, \texttt{road} and \texttt{traffic sign}) with easily-distinguished visual properties. This phenomenon causes imbalanced gradient propagation between easy-to-forget and hard-to-forget old categories, thus exacerbating heterogeneous forgetting speeds among different old classes. When autonomous vehicles \cite{8569992} learn new classes continually, such forgetting heterogeneity aggravates forgetting on those hard-to-forget old classes (\emph{e.g.}, \texttt{traffic sign}) to some extent.

Inspired by the above practical scenario, we investigate that different old classes have significant forgetting heterogeneity in this paper, as shown in Fig.~\ref{fig: motivation}. This heterogeneous forgetting might heavily weaken forgetting compensation on hard-to-forget old classes, as new classes become a subset of old classes consecutively. In summary, the challenges to tackle heterogeneous forgetting lie in two major aspects:

\begin{myitemize}
\item \textbf{Representation Aspect:} 
Some easy-to-forget old classes (\emph{e.g.}, \texttt{car} and \texttt{bus}) with diverse appearances are more difficult for existing CIL methods \cite{Tiwari_2022_CVPR, Yan_2021_CVPR, Dong_2022_CVPR, 10.1007/978-3-030-01219-9_9} to learn intrinsic task-shared representations, and thus are significantly easier to be forgotten than hard-to-forget old classes (\emph{e.g.}, \texttt{road}) with distinctive attributes. Thus, exploring task-shared representations is essential to address heterogeneous forgetting among different old classes.

\item \textbf{Gradient Aspect:} 
To learn complex visual characterizations for easy-to-forget old classes with various modalities and appearances, existing CIL methods \cite{Rebuffi_2017_CVPR, 10.1007/978-3-031-19812-0_7, Wang_ICLR2022} are required to allocate more network architectures for gradient updating. It can result in imbalanced gradient propagation between easy-to-forget and hard-to-forget old categories, thus aggravating forgetting heterogeneity of old classes.

\end{myitemize}

To overcome the above challenges, we propose a novel \underline{\textbf{H}}eterogeneous \underline{\textbf{F}}orgetting \underline{\textbf{C}}ompensation (\textbf{HFC}) model, which is an earlier exploration to address heterogeneous forgetting from both representation and gradient perspectives in the CIL field \cite{Rebuffi_2017_CVPR, 10.1007/978-3-031-19812-0_7}. Specifically, we propose a task-semantic aggregation (TSA) block to alleviate heterogeneous forgetting from representation perspective. It can explore task-shared global representations for each class via aggregating long-range local category information within each task. Meanwhile, to tackle heterogeneous forgetting from gradient perspective, we develop two novel plug-and-play losses: a gradient-balanced forgetting compensation (GFC) loss and a gradient-balanced relation distillation (GRD) loss. The GFC loss can rectify heterogeneous forgetting speeds of easy-to-forget and hard-to-forget old classes, while normalize different learning paces of new categories to achieve gradient-balanced propagation. Besides, the GRD loss can distill heterogeneous relation consistency brought by forgetting heterogeneity among different old classes. Experiments show our model has large improvements on several representative datasets, compared with baseline methods \cite{Rebuffi_2017_CVPR, Liu2023Online}. More importantly, we apply two plug-and-play losses into existing distillation-based CIL methods \cite{Rebuffi_2017_CVPR, 10.1007/978-3-030-58565-5_6, 10.1007/978-3-031-19812-0_7} to significantly improve their performance from gradient aspect. The novel contributions of this paper are presented as follows: 

\begin{myitemize}
\item We develop a novel Heterogeneous Forgetting Compensation (HFC) model to address different forgetting speeds of easy-to-forget and hard-to-forget old classes. To our best knowledge, this paper is the first exploration to tackle forgetting heterogeneity among old categories from representation and gradient aspects in the CIL field. 

\item  We design a task-semantic aggregation (TSA) block to alleviate heterogeneous forgetting from representation aspect. It can explore robust task-shared representations for each class via aggregating local category information.

\item We propose two novel plug-and-play losses: a gradient-balanced forgetting compensation (GFC) loss and a gradient-balanced relation distillation (GRD) loss to surmount forgetting heterogeneity from gradient aspect. They can balance different forgetting of old classes and heterogeneous category-relation consistency to improve performance when applying them into existing methods. 


\end{myitemize}

\section{Related Work}
We discuss some class-incremental learning (CIL) methods \cite{zhao2021mgsvf, DBLP:conf/nips/LiuSS21, Douillard_2022_CVPR}, and divide them into three categories:

\textbf{Knowledge Distillation:} After Li \emph{et al.} \cite{8107520} apply knowledge distillation \cite{44873_Distilling} to continual learning, \cite{Rebuffi_2017_CVPR, NEURIPS2019_83da7c53, 9156766} follow them to transfer past experience from old model to new model. iCaRL \cite{Rebuffi_2017_CVPR} and EEIL \cite{10.1007/978-3-030-01258-8_15} perform knowledge distillation in the output space. \cite{10.1007/978-3-030-58529-7_16} aims to preserve consistent topology feature space for old and new tasks. \cite{douillard2021plop} proposes feature-level knowledge distillation when applying continual learning \cite{DBLP:conf/nips/LiuSS21} into semantic segmentation task \cite{xu2023rssformer, xu2023wave, 9616392_Dong}.

\textbf{Rehearsal Strategy:} For replaying past experience, lots of CIL methods \cite{Wang_ICLR2022, 9815145, tang2022learning} allocate a memory to store exemplars of old classes or synthesize samples via generative adversarial model \cite{NIPS2014_5ca3e9b1, zhang2021causaladv}. For saving large memory overhead, \cite{10.1007/978-3-030-58517-4_41} proposes to store low-dimension features rather than raw samples. \cite{9578918} proposes a ``white box'' framework deviated from rate reduction to preserve past experience.

\textbf{Dynamic Architecture:} Many works \cite{10.5555/3454287.3455512, Yan_2021_CVPR, 9156310} allocate dynamical networks for new classes as the growing of learned classes. Yoon \emph{et al.} \cite{yoon2018lifelong} dynamically expand architecture capacity via selective retraining. \cite{Yan_2021_CVPR, Douillard_2022_CVPR} address the problem of relying on task index when performing dynamic architectures. However, the above-mentioned CIL methods \cite{Rebuffi_2017_CVPR, Hu_2021_CVPR} neglect negative influence of heterogeneous forgetting of old classes on forgetting compensation.

\begin{figure*}[t]
\centering
\includegraphics[trim = 8mm 57mm 9mm 56mm, clip, width=496pt, height=180pt]
{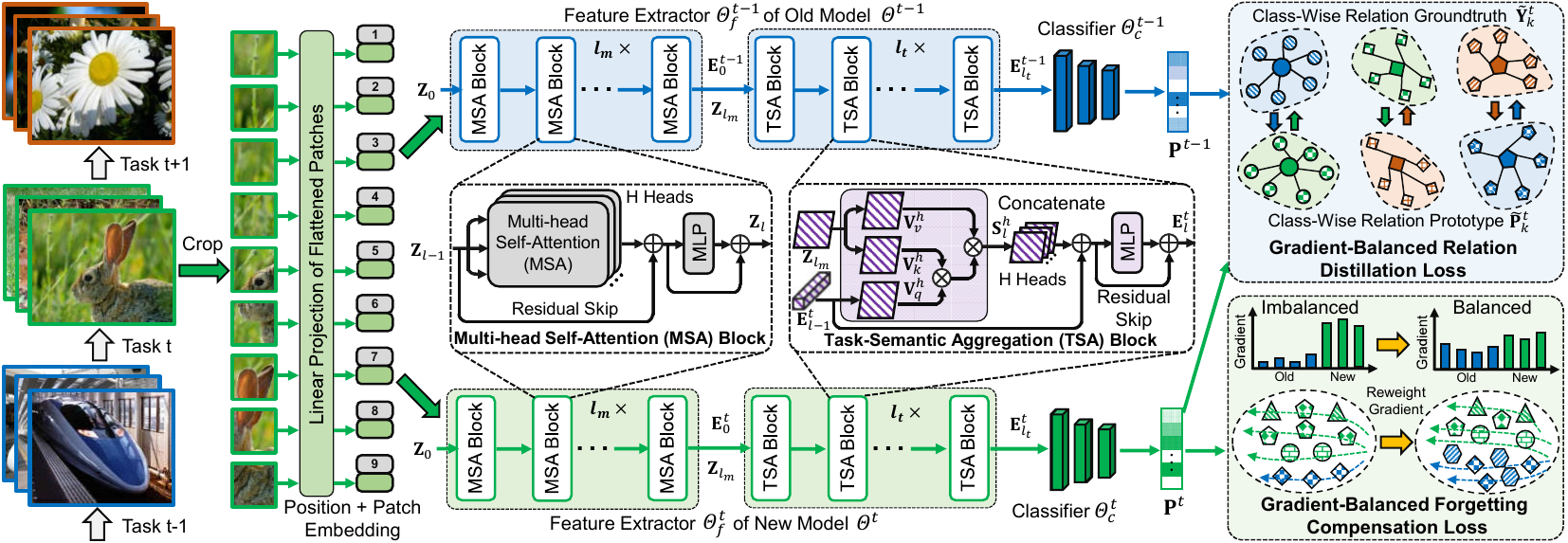}
\vspace{-15pt}
\caption{Illustration of the HFC model. It mainly contains a \textit{task-semantic aggregation (TSA)} block to alleviate heterogeneous forgetting from representation aspect via exploring task-shared global representations, and two novel plug-and-play losses (\emph{i.e.}, a \textit{gradient-balanced forgetting compensation loss} $\mathcal{L}_{\rm{FC}}$ and a \textit{gradient-balanced relation distillation loss} $\mathcal{L}_{\rm{RD}}$) to overcome forgetting heterogeneity from gradient aspect by compensating gradient-imbalanced propagation and heterogeneous class-relation consistency.} 
\label{fig: overview_of_our_model}
\vspace{-10pt}
\end{figure*}

\section{The Proposed Model}\label{sec: proposed_model}

\subsection{Problem Definition and Overview}
\textbf{Problem Definition:} 
Following class-incremental learning (CIL) methods \cite{Rebuffi_2017_CVPR, wu2019large, 10.1007/978-3-030-58565-5_6, Douillard_2022_CVPR, Yan_2021_CVPR}, we set a series of consecutive learning tasks as $\mathcal{T} = \{\mathcal{T}^t\}_{t=1}^T$, where $T$ indicates the total task quantity. The $t$-th task $\mathcal{T}^t=\{\mathbf{x}_i^t, \mathbf{y}_i^t\}_{i=1}^{N^t}$ is composed of $N^t$ pairs of image $\mathbf{x}_i^t$ and one-hot groundtruth $\mathbf{y}_i^t\in\mathcal{Y}^t$, and $\mathcal{Y}^t$ denotes label space of the $t$-th incremental task containing $K^t$ new categories. The label spaces between any two incremental tasks have no overlap: $\mathcal{Y}^t\cap(\cup_{j=1}^{t-1}\mathcal{Y}^j) = \emptyset$. Namely, $K^t$ new classes belonging to the $t$-th task $\mathcal{T}^t$ are different from $K^o=\sum_{i=1}^{t-1}K^i \subset \cup_{j=1}^{t-1}\mathcal{Y}^j$ old classes learned from previous tasks $\{\mathcal{T}^i\}_{i=1}^{t-1}$. As introduced in CIL baselines \cite{Rebuffi_2017_CVPR, wu2019large, Yan_2021_CVPR}, we set a fixed exemplar memory $\mathcal{M}$, and store only few images (\emph{i.e.}, $\frac{|\mathcal{M}|}{K^o}$) for each old class in the $t$-th task $\mathcal{T}^t$, where $\mathcal{M}$ satisfies $\frac{N^t}{K^t} \gg \frac{|\mathcal{M}|}{K^o}$. In the $t$-th task, we aim to identify both $K^t$ new classes and $K^o$ old classes via optimizing the model on $\mathcal{T}^t$ and $\mathcal{M}$.

\textbf{Overview:}
The overview of our HFC model to surmount forgetting heterogeneity of different old categories from both representation and gradient aspects is shown in Fig.~\ref{fig: overview_of_our_model}. Denote the proposed HFC model learned at the $(t\!-\!1)$-th and $t$-th tasks as old and new models (\emph{i.e.}, $\Theta^{t-1}$ and $\Theta^t$), where $\Theta^t$ is inherited from $\Theta^{t-1}$ and only expands the number of output neurons in the classifier to identify $K^t$ new classes. In the $t$-th incremental task, given an image $\mathbf{x}_i^t\in\mathcal{T}^t\cup\mathcal{M}$, we forward it into $\Theta^{t-1}$ and $\Theta^t$ to extract task-shared global representation via task-semantic aggregation (TSA) block, which  alleviates heterogeneous forgetting from representation aspect (Section~\ref{sec: transformer}). After obtaining probabilities $\mathbf{P}^{t-1}(\mathbf{x}_i^t, \Theta^{t-1})\in\mathbb{R}^{K^o}$ and $\mathbf{P}^t(\mathbf{x}_i^t, \Theta^t)\in\mathbb{R}^{K^o+K^t}$ predicted  via $\Theta^{t-1}$ and $\Theta^t$, we develop two plug-and-play losses: a gradient-balanced forgetting compensation (GFC) loss $\mathcal{L}_{\mr{FC}}$ and a gradient-balanced relation distillation (GRD) loss $\mathcal{L}_{\mr{RD}}$ to surmount heterogeneous forgetting from semantic aspect (Sections~\ref{sec: forgetting_compensation} and \ref{sec: relation_distillation}).
$\mathcal{L}_{\mr{FC}}$ and $\mathcal{L}_{\mr{RD}}$ can normalize heterogeneous forgetting of old classes and distill heterogeneous relation consistency between $\Theta^{t-1}$ and $\Theta^t$.

\subsection{Task-Semantic Aggregation Block}\label{sec: transformer}
To explore long-range semantic dependencies across different tasks, Douillard \emph{et al.} \cite{Douillard_2022_CVPR} introduce
Vision Transformer (ViT) \cite{dosovitskiy2020vit, 9710634, He_2022_CVPR} into class-incremental learning (CIL). However, it significantly increases the memory overhead to store task tokens and task-specific decoders, when learning large-scale new classes. Besides, it cannot aggregate local category information to explore task-shared global representations, resulting in heterogeneous forgetting from representation aspect. To tackle these issues, we develop a task-semantic aggregation (TSA) block to extract task-shared global representations, which is effective to tackle heterogeneous forgetting from representation aspect. More importantly, the TSA block cannot dynamically increase memory cost with the arrival of large-scale new categories.

Specifically, as presented in Fig.~\ref{fig: overview_of_our_model}, the proposed model $\Theta^t=\Theta_c^t\circ\Theta_f^t$ at the $t$-th task is composed of a feature extractor $\Theta_f^t$ to extract task-shared global representations, and a task classifier $\Theta_c^t$ to classify $K^o$ old classes and $K^t$ new classes. In the $t$-th task, the network parameters of $\Theta_f^{t-1}$ and $\Theta_c^{t-1}$ in $\Theta^{t-1}$ learned at the $(t\!-\!1)$-th task are frozen to preserve semantic knowledge for old classes. Different from \cite{Douillard_2022_CVPR} that uses vanilla ViT \cite{dosovitskiy2020vit, jaegle2021perceiver, zhang2023dualgats, chen2023vlp} to extract features, we employ $l_m$ multi-head self-attention (MSA) blocks \cite{10.5555/3305381.3305510} and $l_t$ task-semantic aggregation (TSA) block as feature extractor $\Theta_f^t$ to obtain task-shared global representations.




$\bullet$ \textbf{Multi-Head Self-Attention (MSA):} 
Given an image $\mathbf{x}_i^t\in\mathcal{T}^t\cup\mathcal{M}$ in the $t$-th task, following ViT \cite{dosovitskiy2020vit, d2021convit}, we crop it into $N$ patches with equal dimension, flatten these patches and map them into a $D$-dimension feature space via a linear projection to obtain patch embedding $\mathbf{Z}_e\in\mathbb{R}^{N\times D}$. Then we concatenate $\mathbf{Z}_e$ with a class token $\mathbf{Z}_{\mr{cls}}\in\mathbb{R}^{D}$, and perform element-wise sum operation with a position token $\mathbf{Z}_p\in\mathbb{R}^{(N+1)\times D}$ to get $\mathbf{Z}_0 = [\mathbf{Z}_e\oplus\mathbf{Z}_{\mr{cls}}]+\mathbf{Z}_p\in\mathbb{R}^{(N+1)\times D}$, where $\oplus$ is concatenation function. As shown in Fig.~\ref{fig: overview_of_our_model}, $\mathbf{Z}_0$ is fed into $\Theta_f^t$ including $l_m$ MSA blocks to extract long-range semantics across different tasks. For the $l$-th ($l=1, \cdots, l_m$) MSA block, we regard $\mathbf{Z}_{l-1}\in\mathbb{R}^{(N+1)\times D}$ learned from the $(l\!-\!1)$-th MSA block as input, and execute parallel self-attention $H$ times. Thus, the self-attention $\mathbf{A}_{l}^h\in\mathbb{R}^{(N+1) \times d}$ of $\mathbf{Z}_{l-1}$ for the $h$-th ($h=1,\cdots, H$) head is:
\begin{align}
	\label{eq: self_attention}
	\!\! \mathbf{A}_{l}^h =  \sigma(\frac{\mathbf{Z}_{l-1} \mathbf{W}_q^h (\mathbf{Z}_{l-1} \mathbf{W}_k^h)^{\top}}{\sqrt{d}})(\mathbf{Z}_{l-1} \mathbf{W}_v^h), 
\end{align}
where $\mathbf{W}_q^h, \mathbf{W}_k^h, \mathbf{W}_v^h\in\mathbb{R}^{D\times d}$ are query, key and value mapping matrices. $d=\frac{D}{H}$ represents channel dimension of each head and $\sigma$ is the softmax function. Then we concatenate $\{\mathbf{A}_{l}^h\}_{h=1}^H$ along channel dimension and project this concatenated result via a mapping matrix $\mathbf{W}_o\in\mathbb{R}^{D\times D}$ to perform MSA: $\mr{MSA}(\mathbf{Z}_{l-1}) = \mathbf{Z}_{l-1}+[\mathbf{A}_{l}^1 \oplus\mathbf{A}_{l}^2, \cdots, \oplus\mathbf{A}_{l}^H]\mathbf{W}_o$, where $\oplus$ is concatenation function. After obtaining $\mr{MSA}(\mathbf{Z}_{l-1})\in\mathbb{R}^{(N+1)\times D}$ ($D=dH$), we utilize a multi-layer perceptron (MLP) block to obtain output $\mathbf{Z}_l \in\mathbb{R}^{(N+1)\times D}$ of the $l$-th MSA block:
\begin{align}
	\label{eq: output_MHSA}
	\mathbf{Z}_l = \mr{MSA}(\mathbf{Z}_{l-1}) + \mr{MLP}(\mr{MSA}(\mathbf{Z}_{l-1})). 
\end{align}

$\bullet$ \textbf{Task-Semantic Aggregation (TSA):}  
The patch embedding $\mathbf{Z}_{l_m}\in\mathbb{R}^{(N+1)\times D}$ encoded via the $l_m$-th MSA block contains rich long-range semantic dependencies shared across different tasks. $\mathbf{Z}_{l_m}$ is then forwarded into $l_t$ task-semantic aggregation (TSA) blocks to extract task-shared global embedding. The TSA block aggregates local category context from long-range semantic dependencies to global representations shared across all incremental tasks, which is essential to alleviate heterogeneous forgetting from representation aspect. In the $t$-th task, as depicted in Fig.~\ref{fig: overview_of_our_model}, we introduce a learnable task-shared embedding $\mathbf{E}_0^t\in\mathbb{R}^{D}$ that is initialized via $\mathbf{E}_{0}^{t-1}$ learned in the $(t\!-\!1)$-th task. Then we forward $\mathbf{E}_0^t$ along with $\mathbf{Z}_{l_m}$ to $l_t$ TSA blocks. For the $l$-th ($l=1, \cdots, l_t$) TSA block, $\mathbf{E}_{l-1}^t\in\mathbb{R}^D$ learned at $(l-1)$-th TSA block and $\mathbf{Z}_{l_m}$ are used to perform task-semantic attention $H$ times in parallel. Thus, we formulate task-semantic attention $\mathbf{S}_{l}^h\in\mathbb{R}^{d}$ for the $h$-th ($h=1, \cdots, H$) head as: 
\begin{align}
	\label{eq: task_semantic_attention}
	\mathbf{S}_{l}^h =  \sigma(\frac{\mathbf{E}_{l-1}^t \mathbf{V}_q^h (\mathbf{Z}_{l_m} \mathbf{V}_k^h)^{\top}}{\sqrt{d}})(\mathbf{Z}_{l_m}\mathbf{V}_v^h),
\end{align}
where $\mathbf{V}_q^h, \mathbf{V}_k^h, \mathbf{V}_v^h\in\mathbb{R}^{D\times d}$ are projection matrices of query, key and value. After utilizing Eq.~\eqref{eq: task_semantic_attention} to obtain $\{\mathbf{S}_{l}^h\}_{h=1}^H$, we concatenate them along channel dimension and project this fused result via $\mathbf{V}_o\in\mathbb{R}^{D\times D}$ to execute $H$-head task-semantic attention: $\mr{TSA}(\mathbf{E}_{l-1}^t) = [\mathbf{S}_{l}^1\oplus\mathbf{S}_{l}^2, \cdots, \oplus\mathbf{S}_{l}^H]\mathbf{V}_o\in\mathbb{R}^{D}$. $\mr{TSA}(\mathbf{E}_{l-1}^t)$ is encoded via a MLP block to get output $\mathbf{E}_l^t\in\mathbb{R}^{D}$ of the $l$-th TSA block: 
\begin{align}
	\label{eq: output_TSP}
	\mathbf{E}_l^t = \mr{TSA}(\mathbf{E}_{l-1}^t) + \mr{MLP}(\mr{TSA}(\mathbf{E}_{l-1}^t)).   
\end{align}

As aforementioned, we obtain the task-shared global representation $\mathbf{E}_{l_t}^t \in\mathbb{R}^{D}$ via the $l_t$-th TSA block in feature extractor $\Theta_f^t$. Obviously, the proposed TSA blocks can aggregate local category context of the $t$-th task from long-range semantic dependencies to task-shared global representation $\mathbf{E}_{l_t}^t$. It is essential to alleviate heterogeneous forgetting among different old classes from representation aspect, by exploring task-shared semantic context across different tasks. Note that we employ layer norm before TSA, MSA and MLP blocks, while we omit it for simplicity in this paper.

$\bullet$ \textbf{Task Classifier:}
The task-shared global representation $\mathbf{E}_{l_t}^t$ obtained via $\Theta_f^t$ is fed into task classifier $\Theta_c^t$ to predict $K^o$ old classes and $K^t$ new classes. Given a mini-batch $\{\mf{x}_i^t, \mf{y}_i^t\}_{i=1}^b\in\mathcal{T}^t\cup\mathcal{M}$, the classification loss $\mathcal{L}_{\mr{CE}}$ is:
\begin{align}
	\mathcal{L}_{\mr{CE}} =\frac{1}{b}\sum_{i=1}^b \mathcal{D}_{\mr{CE}}(\mathbf{P}^t(\mathbf{x}_{i}^t, \Theta^t), \mathbf{y}_{i}^t),
	\label{eq: classification_loss} 
\end{align}
where $\mathbf{P}^t(\mathbf{x}_{i}^t, \Theta^t)\in\mathbb{R}^{K^o+K^t}$ is softmax probability predicted by $\Theta^t$ in the $t$-th learning task. $\mathcal{D}_{\mr{CE}}(\cdot, \cdot)$ indicates traditional cross-entropy loss, and $b$ is the batch size.

However, the severe class imbalance (\emph{i.e.}, $\frac{N^t}{K^t} \gg \frac{|\mathcal{M}|}{K^o}$) among old and new categories in the $t$-th task enforces the prediction $\mathbf{P}^t(\mathbf{x}_{i}^t, \Theta^t)$ in Eq.~\eqref{eq: classification_loss} to suffer from catastrophic forgetting \cite{Rebuffi_2017_CVPR} on old categories. Moreover, the gradient optimization of Eq.~\eqref{eq: classification_loss} neglects heterogeneous forgetting speeds of easy-to-forget old classes with various appearances and hard-to-forget old classes with easily-distinguished attributes. To address heterogeneous forgetting from gradient aspect, we propose two novel plug-and-play losses: a gradient-balanced forgetting compensation (GFC) loss $\mathcal{L}_{\mr{FC}}$ in Section~\ref{sec: forgetting_compensation} and a gradient-balanced relation distillation (GRD) loss $\mathcal{L}_{\mr{RD}}$ in Section~\ref{sec: relation_distillation}.

\subsection{Gradient-Balanced Forgetting Compensation}\label{sec: forgetting_compensation} 
As incremental tasks arrive consecutively, easy-to-forget and hard-to-forget old categories change dynamically, thus aggravating difficulty to tackle heterogeneous forgetting from gradient aspect. In light of this, we develop a gradient-balanced forgetting compensation (GFC) loss $\mathcal{L}_{\mr{FC}}$. It adaptively balances large forgetting heterogeneity of old classes via performing balanced gradient propagation. Specifically, we perform task-adaptive gradient normalization for different classes, and then reweight classification loss $\mathcal{L}_{\mr{CE}}$ in Eq.~\eqref{eq: classification_loss} to compensate imbalanced gradients. For the $t$-th incremental task, as claimed in \cite{wang2021addressing, dong2023federated_FISS}, the gradient $\Gamma_i^t$ of an given image $(\mf{x}_{i}^t, \mf{y}_i^t)\subset\mathcal{T}^t\cup\mathcal{M}$ with respect to the $k$-th ($k=\arg\max \mathbf{y}_i^t$) neuron $\mathcal{N}_k^t$ of classifier $\Theta_c^t$ in $\Theta^t$ is:
\begin{align}
	\Gamma_i^t= \frac{\partial\mathcal{D}_{\mr{CE}}(\mathbf{P}^t(\mathbf{x}_{i}^t, \Theta^t), \mathbf{y}_{i}^t)}{\partial\mathcal{N}_k^t} = \mathbf{P}^t(\mf{x}_{i}^t, \Theta^t)_k - 1,
	\label{eq: gradient_value}
\end{align}
where $\mathbf{P}^t(\mf{x}_{i}^t, \Theta^t)_k$ denotes probability of the $k$-th class.

To compensate heterogeneous forgetting of easy-to-forget and hard-to-forget old categories, we compute task-adaptive gradient means for different tasks. Consequently, given a mini-batch $\{\mathbf{x}_{i}^t,\mathbf{y}_{i}^t\}_{i=1}^b$ in the $t$-th learning task, the task-adaptive gradient mean $\Gamma_\eta$ for the categories learned in the $\eta$-th ($1\leq\eta\leq t$) incremental task is formulated as:
\begin{align}
	\Gamma_\eta = \frac{1}{\sum_{i=1}^b \mathbb{I}_{\mf{y}_{i}^t\in\mathcal{Y}^\eta}} \sum\nolimits_{i=1}^b |\Gamma_i^t|\cdot \mathbb{I}_{\mf{y}_i^t\in\mathcal{Y}^\eta}.
	\label{eq: task_specific_gradient}
\end{align}
The forgetting heterogeneity of old categories learned from $t\!-\!1$ old tasks and learning speeds of new categories in the $t$-th task can be effectively measured via $\{\Gamma_\eta\}_{\eta=1}^t$ in Eq.~\eqref{eq: task_specific_gradient}.

When easy-to-forget and hard-to-forget old classes are varied dynamically as incremental tasks, some noisy predictions on easy-to-forget old classes may result in large deviation to measure their forgetting heterogeneity \cite{tang2022virtual, NEURIPS2022_f0e91b13}. To tackle this issue, $\{\Gamma_\eta\}_{\eta=1}^t$ in Eq.~\eqref{eq: task_specific_gradient} are expected to be sharper and more distinguishable adaptively \cite{dong2023federated_FISS}, as learning tasks arrive continually. Thus, we rewrite Eq.~\eqref{eq: task_specific_gradient} as follows:
\begin{align}
	\Gamma_{\eta}^s = \frac{\sum\nolimits_{i=1}^b \log(|\Gamma_{i}^t|^{\frac{K^o}{K^o+K^t}}+1)\cdot \mathbb{I}_{\mf{y}_{i}^t\in\mathcal{Y}^{\eta}}}{\sum_{i=1}^b \mathbb{I}_{\mf{y}_{i}^t\in\mathcal{Y}^\eta}}. 
	\label{eq: task_specific_gradient_sharper}
\end{align}
The sharper task-adaptive gradient means $\{\Gamma_{\eta}^s\}_{\eta=1}^t$ computed via Eq.~\eqref{eq: task_specific_gradient_sharper} are employed to reweight classification loss $\mathcal{L}_{\mr{CE}}$ in Eq.~\eqref{eq: classification_loss}. Then the gradient-balanced forgetting compensation (GFC) loss $\mathcal{L}_{\mr{FC}}$ is written as follows:
\begin{align}
	\!\!\! \mathcal{L}_{\mr{FC}}\!=\! \frac{1}{b}\sum_{i=1}^b \frac{\log(|\Gamma_i^t|^{\frac{K^o}{K^o+K^t}}\!\!+\!\!1)}{\sum_{\eta=1}^t \Gamma_\eta^s\cdot\mathbb{I}_{\mf{y}_{i}^t\in\mathcal{Y}^{\eta}}}
	\mathcal{D}_{\mr{CE}}(\mathbf{P}^t(\mathbf{x}_{i}^t, \Theta^t),  \mathbf{y}_{i}^t).
	\label{eq: gradient_balanced_compensation}
\end{align}
Obviously, $\mathcal{L}_{\mr{FC}}$ in Eq.~\eqref{eq: gradient_balanced_compensation} encourages the optimization of $\Theta^t$ to perform gradient-balanced propagation for easy-to-forget, hard-to-forget old classes and new classes learned in different tasks. It tackles heterogeneous forgetting on old classes from gradient aspect, via normalizing forgetting paces of different old classes adaptively with $\{\Gamma_\eta^s\}_{\eta=1}^t$.

\subsection{Gradient-Balanced Relation Distillation}\label{sec: relation_distillation}
The relationships among old and new categories remain constant in semantic space, regardless of availability of old-class samples. In light of this, distilling inter-class relations from old model $\Theta^{t-1}$ to new model $\Theta^t$ can address forgetting on old classes \cite{10.1007/978-3-030-01258-8_15, Hou_2019_CVPR}. However, most knowledge distillation methods used in CIL \cite{Rebuffi_2017_CVPR, Ahn_2021_ICCV, wu2019large} only utilize prediction of an individual sample to perform semantic consistency of old classes between $\Theta^{t-1}$ and $\Theta^t$. They neglect underlying relations among old and new categories to tackle forgetting, and heavily suffer from noisy predictions on easy-to-forget old classes when distilling knowledge via an individual sample. Moreover, different forgetting speeds of old classes may result in heterogeneous class-relation consistency between $\Theta^{t-1}$ and $\Theta^t$. It enforces gradient optimization of $\Theta^t$ to bias towards some specific relation distillations related to new classes, causing forgetting heterogeneity on old categories.

To overcome the above issues, we design a gradient-balanced relation distillation (GRD) loss $\mathcal{L}_{\mr{RD}}$ to distill relations among old and new classes from $\Theta^{t-1}$ to $\Theta^t$. As introduced in Fig.~\ref{fig: overview_of_our_model}, it can address heterogeneous forgetting from gradient aspect via rectifying imbalanced gradient propagation caused by heterogeneous class-relation consistency. Specifically, we construct class-wise relation prototype rather than a prediction of single sample to perform relation distillation, which can alleviate negative effect of noisy relation consistency. Then the task-adaptive gradient means $\{\Gamma_\eta^s\}_{\eta=1}^t$ in Eq.~\eqref{eq: task_specific_gradient_sharper} are employed to reweight heterogeneous class-relation distillation.

\begin{table*}[t]
\centering
\setlength{\tabcolsep}{1.3mm}
\caption{Results on CIFAR-100 \cite{krizhevsky2009learning}. When $\mathcal{M}=2000$, we set $T\!=\!\{5, 10, 20, 25, 50\}$ for $\mathcal{B}\!=\!0\%$, and $T\!=\!\{5, 10, 25, 50\}$ for $\mathcal{B}\!=\!50\%$.  }
\scalebox{0.82}{
\begin{tabular}{l|cc|ccccc|cl|cccc|cl}
	\toprule
\makecell[c]{\multirow{2}{*}{Comparison Methods}} & \multirow{2}{*}{Backbone} & \multirow{2}{*}{\#Params} & \multicolumn{7}{c|}{$\mathcal{B}=0\%$} & \multicolumn{6}{c}{$\mathcal{B}=50\%$}  \\
	  &  &  & 5 & 10 & 20 & 25 & 50 & Avg. & Imp. & 5 & 10 & 25 & 50 & Avg. & Imp. \\
	\midrule
 iCaRL \cite{Rebuffi_2017_CVPR} (CVPR'2017) & ViT-Base & 85.10M  &83.5 &80.6 &78.5 &77.6 &76.4 &79.3 &$\Uparrow$5.5 &80.6 &78.6 &75.4 &71.0 &76.4 &$\Uparrow$2.9  \\
	BiC \cite{wu2019large} (CVPR'2019) & ViT-Base & 85.10M &83.5 &82.5 &78.7 &80.3 &78.4 &80.7 &$\Uparrow$4.2 &78.2 &75.7 &74.2 &71.9 &75.0 &$\Uparrow$4.3\\
	PODNet \cite{10.1007/978-3-030-58565-5_6} (ECCV'2020) & ViT-Base & 85.10M &75.3 &71.8 &66.5 &64.9 &63.0 &68.3 &$\Uparrow$16.6 &72.8 &73.0 &74.2 &74.9 &73.7 &$\Uparrow$5.6  \\
	SS-IL \cite{Ahn_2021_ICCV} (ICCV'2021) & ViT-Base& 85.10M &83.6 &82.0 &79.1 &79.1 &77.3 &80.2 &$\Uparrow$4.6  &75.8 &74.8 &72.9 &72.0 &73.9 &$\Uparrow$5.4  \\
 PODNet \cite{10.1007/978-3-030-58565-5_6} + CSCCT \cite{10.1007/978-3-031-19812-0_7} (ECCV'2022) & ViT-Base & 85.10M &75.2 &72.0 &66.1 &65.2 &63.8 &68.5 &$\Uparrow$16.4 &72.7 &72.8 &74.0 &74.7 &73.5 &$\Uparrow$5.8\\
 FOSTER \cite{10.1007/978-3-031-19806-9_23} (ECCV'2022) & ViT-Base & 85.10M &84.7 &83.5 &79.6 &78.5 &77.9 &80.8 &$\Uparrow$4.1 &78.4 &77.3 &76.2 &74.0 &76.5 &$\Uparrow$2.8 \\
        AFC \cite{Kang_2022_CVPR} (CVPR'2022) &ViT-Base & 85.10M &74.6 &75.5 &70.3 &67.4 &64.7 &70.5 &$\Uparrow$14.4 &70.8 &69.4 &68.9 &69.9 &69.8 &$\Uparrow$9.5 \\
	DyTox \cite{Douillard_2022_CVPR} (CVPR'2022) & ViT-Base & 85.10M &85.5 &85.8 &81.3 &80.8 &77.4 &82.2 &$\Uparrow$2.7 &82.5 &81.0 &77.3 &74.5 &78.8 &$\Uparrow$0.5 \\
        \midrule
        \rowcolor{lightgray}
        \textbf{HFC} (\textbf{Ours}) & ViT-Base & 85.10M  &\textcolor{deepred}{\textbf{86.4}} &\textcolor{deepred}{\textbf{86.3}} &\textcolor{deepred}{\textbf{85.5}} &\textcolor{deepred}{\textbf{85.0}} &\textcolor{deepred}{\textbf{81.1}} &\textcolor{deepred}{\textbf{84.9}} &\textbf{$\mathrm{-}$} &\textcolor{deepred}{\textbf{82.9}} &\textcolor{deepred}{\textbf{81.0}} &\textcolor{deepred}{\textbf{77.8}} &\textcolor{deepred}{\textbf{75.5}} &\textcolor{deepred}{\textbf{79.3}} &\textbf{$\mathrm{-}$} \\
        \midrule
        Upper Bound & ViT-Base & 85.10M &94.2 &94.2 &94.2 &94.2 &94.2 & 94.2 &\textbf{$\mathrm{-}$} &94.2 &94.2 &94.2 &94.2 & 94.2 &\textbf{$\mathrm{-}$}\\
 \bottomrule
\end{tabular}}
\label{tab: comparison_exp_cifar100}
\vspace{-10pt}
\end{table*}

\begin{table*}[t]
\centering
\setlength{\tabcolsep}{1.65mm}
\caption{Results on ImageNet-100 \cite{5206848}. When $\mathcal{M}=2000$, we set $T=\{5, 10, 20, 25\}$ for $\mathcal{B}=0\%$, and $T=\{5, 10, 25, 50\}$ for $\mathcal{B}=50\%$.  }
\scalebox{0.82}{
\begin{tabular}{l|cc|cccc|cl|cccc|cl}
	\toprule
\makecell[c]{\multirow{2}{*}{Comparison Methods}} & \multirow{2}{*}{Backbone} & \multirow{2}{*}{\#Params} & \multicolumn{6}{c|}{$\mathcal{B}=0\%$} & \multicolumn{6}{c}{$\mathcal{B}=50\%$}  \\
	  &  &  & 5 & 10 & 20 & 25 & Avg. & Imp. & 5 & 10 & 25 & 50 & Avg. & Imp. \\
	\midrule
 iCaRL \cite{Rebuffi_2017_CVPR} (CVPR'2017) & ViT-Base & 85.10M &84.1 &82.4 &80.2 &78.6 &81.3 &$\Uparrow$3.2  &79.5 &78.8 &76.6 &73.9 &77.2 &$\Uparrow$3.4\\
	BiC \cite{wu2019large} (CVPR'2019) & ViT-Base & 85.10M &81.4 &80.2 &79.2 &78.5 &79.8 &$\Uparrow$4.7 &73.9 &74.4 &73.4 &72.2 &73.5 &$\Uparrow$7.1\\
	PODNet \cite{10.1007/978-3-030-58565-5_6} (ECCV'2020) & ViT-Base & 85.10M &82.4 &80.1 &79.5 &78.3 &80.1 &$\Uparrow$4.5 &68.4 &71.3 &75.4 &77.6 &73.2 &$\Uparrow$7.4 \\
	SS-IL \cite{Ahn_2021_ICCV} (ICCV'2021) & ViT-Base& 85.10M &83.0 &81.7 &80.9 &79.4 &81.2 &$\Uparrow$3.3  &75.8 &74.8 &72.9 &72.0 &73.9  &$\Uparrow$6.7 \\
 PODNet \cite{10.1007/978-3-030-58565-5_6} + CSCCT \cite{10.1007/978-3-031-19812-0_7} (ECCV'2022) & ViT-Base & 85.10M &82.3 &81.0 &79.3 &79.0 &80.4 &$\Uparrow$4.1 &68.3 &71.2 &75.0 &76.9 &72.8 &$\Uparrow$7.8 \\
 FOSTER \cite{10.1007/978-3-031-19806-9_23} (ECCV'2022) & ViT-Base & 85.10M &85.6 &84.0 &83.8 &\textcolor{deepred}{\textbf{83.1}} &84.1 &$\Uparrow$0.4 &81.0 &80.5 &80.0 &\textcolor{deepred}{\textbf{79.8}} &80.3 &$\Uparrow$0.3\\
        AFC \cite{Kang_2022_CVPR} (CVPR'2022) & ViT-Base & 85.10M &83.2 &81.2 &80.5 &79.4 &81.1 &$\Uparrow$3.5 &77.1 &78.4 &78.9 &78.1 &78.1&$\Uparrow$2.5 \\
	DyTox \cite{Douillard_2022_CVPR} (CVPR'2022) & ViT-Base & 85.10M &85.1 &83.4 &80.0 &80.3 &82.2 &$\Uparrow$2.3 &77.4 &78.4 &79.7 &79.4 &78.7 &$\Uparrow$1.9\\ 
 
        \midrule
        \rowcolor{lightgray}
        \textbf{HFC} (\textbf{Ours}) & ViT-Base & 85.10M &\textcolor{deepred}{\textbf{86.1}} &\textcolor{deepred}{\textbf{85.4}} &\textcolor{deepred}{\textbf{84.0}} &82.6 &\textcolor{deepred}{\textbf{84.5}} &\textbf{$\mathrm{-}$} &\textcolor{deepred}{\textbf{81.5}} &\textcolor{deepred}{\textbf{81.1}}  &\textcolor{deepred}{\textbf{80.3}}  &79.5 &\textcolor{deepred}{\textbf{80.6}} &\textbf{$\mathrm{-}$} \\
        \midrule
        Upper Bound & ViT-Base & 85.10M &86.3 &86.3 &86.3 &86.3 &86.3 &\textbf{$\mathrm{-}$} &86.3 &86.3 &86.3 &86.3 &86.3 &\textbf{$\mathrm{-}$} \\
 \bottomrule
 
\end{tabular}}
\label{tab: comparison_exp_image100}
\vspace{-10pt}
\end{table*}

\begin{table}[t]
\centering
\setlength{\tabcolsep}{1.3mm}
\caption{Results on ImageNet-1000 \cite{5206848}. When $\mathcal{M}=20000$, we set $T=\{5, 10\}$ for $\mathcal{B}=0\%$.  }
\scalebox{0.62}{
\begin{tabular}{l|cc|cc|cl}
	\toprule
	Comparison Methods & Backbone & \#Params & 5 & 10 & Avg. & Imp. \\
	\midrule
 iCaRL \cite{Rebuffi_2017_CVPR} (CVPR'2017) & ViT-Base & 85.10M &75.4 &72.4 &73.9 &$\Uparrow$3.5  \\
	BiC \cite{wu2019large} (CVPR'2019) & ViT-Base & 85.10M &68.9 &66.0 &67.5 &$\Uparrow$10.0  \\
	PODNet \cite{10.1007/978-3-030-58565-5_6} (ECCV'2020) & ViT-Base & 85.10M &68.4 &71.3 &69.8 &$\Uparrow$7.6   \\
	SS-IL \cite{Ahn_2021_ICCV} (ICCV'2021) & ViT-Base& 85.10M &75.8 &74.8 &75.3 &$\Uparrow$2.2    \\
 PODNet \cite{10.1007/978-3-030-58565-5_6} + CSCCT \cite{10.1007/978-3-031-19812-0_7} (ECCV'2022) & ViT-Base & 85.10M &68.4 &63.2 &65.8 &$\Uparrow$11.6 \\
 FOSTER \cite{10.1007/978-3-031-19806-9_23} (ECCV'2022) & ViT-Base & 85.10M &76.9 &74.3 &75.6 &$\Uparrow$1.9 \\
        AFC \cite{Kang_2022_CVPR} (CVPR'2022) & ViT-Base & 85.10M  &74.2 &71.4 &72.8 &$\Uparrow$4.6  \\
	DyTox \cite{Douillard_2022_CVPR} (CVPR'2022) & ViT-Base & 85.10M &78.1 &74.7 &76.4 &$\Uparrow$1.0 \\ 
        \midrule
        \rowcolor{lightgray}
        \textbf{HFC} (\textbf{Ours}) & ViT-Base & 85.10M &\textcolor{deepred}{\textbf{78.5}} &\textcolor{deepred}{\textbf{76.4}} &\textcolor{deepred}{\textbf{77.5}} &\textbf{$\mathrm{-}$}  \\
        \midrule
        Upper Bound & ViT-Base & 85.10M &86.3 &86.3 & 86.3 &\textbf{$\mathrm{-}$}   \\
 \bottomrule
 
\end{tabular}}
\label{tab: comparison_exp_image1000}
\vspace{-10pt}
\end{table}

A mini-batch $\{\mf{x}_{i}^t, \mf{y}_{i}^t\}_{i=1}^b\subset \mathcal{T}^t\cup\mathcal{M}$ is fed into $\Theta^{t-1}$ and $\Theta^t$ to obtain softmax probabilities $\mathbf{P}^{t-1}(\mathbf{x}_{i}^t, \Theta^{t-1})\in\mathbb{R}^{K^o}$ of old classes and $\mathbf{P}^t(\mathbf{x}_{i}^t, \Theta^t)\in\mathbb{R}^{K^o+K^t}$ of old and new classes, as shown in Fig.~\ref{fig: overview_of_our_model}. The first $K^o$ dimensions of $\mathbf{y}_{i}^t\in\mathbb{R}^{K^o+K^t}$ are replaced with $\mathbf{P}^{t-1}(\mathbf{x}_{i}^t, \Theta^{t-1})$ to get relation groundtruth $\mathbf{Y}^t(\mathbf{x}_{i}^t, \Theta^{t-1})\in\mathbb{R}^{K^o+K^t}$ that implies underlying relations among old and new classes. To tackle noisy class relations, we construct class-wise relation prototype $\tilde{\mathbf{P}}_k^t$ and its relation groundtruth $\tilde{\mathbf{Y}}_k^t$ for the $k$-th class:
\begin{align}
	 \tilde{\mathbf{P}}_k^t &= \frac{1}{\Delta_k}\sum_{i=1}^b \mathbf{P}^t(\mathbf{x}_{i}^t, \Theta^t)\cdot \mathbb{I}_{\arg\max\mf{y}_{i}^t=k}, \\
	 \tilde{\mathbf{Y}}_k^t &= \frac{1}{\Delta_k}\sum_{i=1}^b \mathbf{Y}^t(\mathbf{x}_{i}^t, \Theta^{t-1})\cdot \mathbb{I}_{\arg\max\mf{y}_{i}^t=k}.
	\label{eq: relation_prototype}
\end{align}
where $\Delta_k \!=\! \sum_i^b \mathbb{I}_{\arg\max\mf{y}_{i}^t=k}$. Then we formulate category-wise gradient mean $\Gamma_k^s$ for the $k$-th category as follows:
\begin{align}
	\Gamma_k^s = \frac{\sum\nolimits_{i=1}^b \log(|\Gamma_{i}^t|^{\frac{K^o}{K^o+K^t}}+1)\cdot \mathbb{I}_{\arg\max\mf{y}_{i}^t=k}}{\sum_{i=1}^b \mathbb{I}_{\arg\max\mf{y}_{i}^t=k}}.
	\label{eq: category_specific_gradient}
\end{align}

Consequently, we use $\Gamma_k^s$ to reweight heterogeneous relation distillation, and express the proposed $\mathcal{L}_{\mr{RD}}$ as follows:
\begin{align}	
	\mathcal{L}_{\mr{RD}} = \frac{1}{K^o\!+\!K^t}\sum_{k=1}^{K^o+K^t} \frac{\Gamma_k^s \cdot \mathcal{D}_{\mr{KL}}(\tilde{\mathbf{P}}_k^t, \tilde{\mathbf{Y}}_k^t) }{\sum_{\eta=1}^t \Gamma_\eta^s\cdot\mathbb{I}_{k\in\mathcal{Y}^{\eta}}},
	\label{eq: class_relation_distillation}
\end{align}
where $\mathcal{D}_{\mr{KL}}(\cdot||\cdot)$ indicates the Kullback-Leibler divergence.

\begin{table*}[t]
\centering
\setlength{\tabcolsep}{1.2mm}
\caption{Performance on CIFAR-100 \cite{krizhevsky2009learning} ($T=\{5, 10, 20, 25, 50\}$) and ImageNet-100 \cite{5206848} ($T=\{5, 10, 20, 25\}$) when we apply \textbf{Ours$^\ddag$} into existing distillation-based CIL methods and set $\mathcal{M}=2000, \mathcal{B}=0\%$. \textbf{Ours$^\ddag$} denotes the proposed plug-and-play losses $\mathcal{L}_{\rm{FC}}$ and $\mathcal{L}_{\rm{RD}}$. }
\scalebox{0.782}{
\begin{tabular}{l|cc|ccccc|c|cc|cccc|c}
	\toprule
\makecell[c]{\multirow{2}{*}{Comparison Methods}} & \multicolumn{8}{c|}{CIFAR-100 Dataset \cite{krizhevsky2009learning}} & \multicolumn{7}{c}{ImageNet-100 Dataset \cite{5206848}}  \\
	  & Backbone & \#Params & 5 & 10 & 20 & 25 & 50 & Avg. & Backbone & \#Params & 5 & 10 & 20 & 25 & Avg.  \\
	\midrule
iCaRL \cite{Rebuffi_2017_CVPR} (CVPR'2017) & ResNet-32 & 0.46M &66.4 &63.9 &53.1 &50.2 &39.1 &54.5 & ResNet-18 & 11.22M &76.4 &69.5 &60.6 &57.9 &66.1  \\
        \rowcolor{lightgray}
        iCaRL \cite{Rebuffi_2017_CVPR} + \textbf{Ours}$^\ddag$ & ResNet-32 & 0.46M &\textcolor{deepred}{\textbf{68.0}} &\textcolor{deepred}{\textbf{65.7}} &\textcolor{deepred}{\textbf{54.2}} &\textcolor{deepred}{\textbf{51.4}} &\textcolor{deepred}{\textbf{40.8}} &\textcolor{deepred}{\textbf{56.0}}& ResNet-18 & 11.22M &\textcolor{deepred}{\textbf{76.8}} &\textcolor{deepred}{\textbf{70.3}} &\textcolor{deepred}{\textbf{61.7}} &\textcolor{deepred}{\textbf{59.2}} &\textcolor{deepred}{\textbf{67.0}} \\
        \midrule
	BiC \cite{wu2019large} (CVPR'2019) & ResNet-32 & 0.46M &55.3 &50.8 &48.3 &46.4 &39.3 &48.0 & ResNet-18 & 11.22M &61.5 &54.3 &46.7 &44.5 &51.8  \\
  \rowcolor{lightgray}
        BiC \cite{wu2019large} + \textbf{Ours}$^\ddag$ & ResNet-32 & 0.46M &\textcolor{deepred}{\textbf{63.2}} &\textcolor{deepred}{\textbf{53.8}} &\textcolor{deepred}{\textbf{52.4}} &\textcolor{deepred}{\textbf{49.3}} &\textcolor{deepred}{\textbf{41.5}} &\textcolor{deepred}{\textbf{52.0}} & ResNet-18 & 11.22M &\textcolor{deepred}{\textbf{63.2}} &\textcolor{deepred}{\textbf{57.0}} &\textcolor{deepred}{\textbf{50.3}} &\textcolor{deepred}{\textbf{48.4}} &\textcolor{deepred}{\textbf{54.7}}   \\
        \midrule
	PODNet \cite{10.1007/978-3-030-58565-5_6} (ECCV'2020) & ResNet-32 & 0.46M &63.8 &54.5 &48.1 &44.4 &39.7 &50.1 & ResNet-18 & 11.22M &74.9 &66.5 &55.8 &51.7 &62.2    \\
 \rowcolor{lightgray}
        PODNet \cite{10.1007/978-3-030-58565-5_6} + \textbf{Ours}$^\ddag$ & ResNet-32 & 0.46M &\textcolor{deepred}{\textbf{65.8}} &\textcolor{deepred}{\textbf{60.0}} &\textcolor{deepred}{\textbf{51.6}} &\textcolor{deepred}{\textbf{48.3}} &\textcolor{deepred}{\textbf{44.7}} &\textcolor{deepred}{\textbf{54.1}} & ResNet-18 & 11.22M &\textcolor{deepred}{\textbf{76.5}} &\textcolor{deepred}{\textbf{69.3}} &\textcolor{deepred}{\textbf{59.4}} &\textcolor{deepred}{\textbf{55.8}} &\textcolor{deepred}{\textbf{65.2}}  \\
        \midrule
	SS-IL \cite{Ahn_2021_ICCV} (ICCV'2021) & ResNet-32 & 0.46M &65.1 &59.0 &49.0 &48.6 &40.7 &52.5 & ResNet-18 & 11.22M &60.9 &53.5 &39.3 &35.8 &47.4  \\
 \rowcolor{lightgray}
 SS-IL \cite{Ahn_2021_ICCV} + \textbf{Ours}$^\ddag$ & ResNet-32 & 0.46M &\textcolor{deepred}{\textbf{67.7}} &\textcolor{deepred}{\textbf{64.9}} &\textcolor{deepred}{\textbf{51.6}} &\textcolor{deepred}{\textbf{51.2}} &\textcolor{deepred}{\textbf{42.5}} &\textcolor{deepred}{\textbf{55.6}} & ResNet-18 & 11.22M &\textcolor{deepred}{\textbf{75.2}} &\textcolor{deepred}{\textbf{67.5}} &\textcolor{deepred}{\textbf{57.9}} &\textcolor{deepred}{\textbf{56.0}} &\textcolor{deepred}{\textbf{64.2}}  \\
 \midrule
  PODNet \cite{10.1007/978-3-030-58565-5_6} + CSCCT \cite{10.1007/978-3-031-19812-0_7} (ECCV'2022) & ResNet-32 & 0.46M  &63.7 &54.3 &47.8 &44.0 &39.0 &49.8 & ResNet-18 & 11.22M &74.8 &66.3 &55.4 &51.3 &62.0   \\
   \rowcolor{lightgray}
   PODNet \cite{10.1007/978-3-030-58565-5_6} + CSCCT \cite{10.1007/978-3-031-19812-0_7} + \textbf{Ours}$^\ddag$ &  ResNet-32 & 0.46M &\textcolor{deepred}{\textbf{66.7}} &\textcolor{deepred}{\textbf{58.5}} &\textcolor{deepred}{\textbf{50.5}} &\textcolor{deepred}{\textbf{46.9}} &\textcolor{deepred}{\textbf{43.3}} &\textcolor{deepred}{\textbf{53.2}} & ResNet-18 & 11.22M &\textcolor{deepred}{\textbf{76.4}} &\textcolor{deepred}{\textbf{68.7}} &\textcolor{deepred}{\textbf{58.9}} &\textcolor{deepred}{\textbf{55.1}} &\textcolor{deepred}{\textbf{64.8}}  \\
   \midrule
 FOSTER \cite{10.1007/978-3-031-19806-9_23} (ECCV'2022) & ResNet-32 & 0.46M &72.0 &68.7 &67.1 &66.0 &62.6 &67.3  & ResNet-18 & 11.22M &74.9 &70.9 &67.2 &65.7 &69.7  \\
 \rowcolor{lightgray}
   FOSTER \cite{10.1007/978-3-031-19806-9_23} + \textbf{Ours}$^\ddag$ & ResNet-32 & 0.46M &\textcolor{deepred}{\textbf{72.2}} &\textcolor{deepred}{\textbf{69.9}} &\textcolor{deepred}{\textbf{67.4}} &\textcolor{deepred}{\textbf{66.7}} &\textcolor{deepred}{\textbf{62.8}} &\textcolor{deepred}{\textbf{67.8}} & ResNet-18 & 11.22M &\textcolor{deepred}{\textbf{75.0}} &\textcolor{deepred}{\textbf{71.3}} &\textcolor{deepred}{\textbf{67.5}} &\textcolor{deepred}{\textbf{66.2}} &\textcolor{deepred}{\textbf{70.0}}   \\
        \midrule
        AFC \cite{Kang_2022_CVPR} (CVPR'2022) & ResNet-32 & 0.46M &65.6 &57.1 &50.9 &48.1 &42.0 &52.7 & ResNet-18 & 11.22M &76.3 &69.7 &61.1 &58.1 &66.3   \\
        \rowcolor{lightgray}
   AFC \cite{Kang_2022_CVPR} + \textbf{Ours}$^\ddag$ & ResNet-32 & 0.46M &\textcolor{deepred}{\textbf{66.2}} &\textcolor{deepred}{\textbf{63.8}} &\textcolor{deepred}{\textbf{53.2}} &\textcolor{deepred}{\textbf{49.7}} &\textcolor{deepred}{\textbf{44.1}} &\textcolor{deepred}{\textbf{55.4}}& ResNet-18 & 11.22M &\textcolor{deepred}{\textbf{77.4}} &\textcolor{deepred}{\textbf{71.2}} &\textcolor{deepred}{\textbf{63.3}} &\textcolor{deepred}{\textbf{67.4}} &\textcolor{deepred}{\textbf{69.8}}  \\
   \midrule
   Upper Bound & ResNet-32 & 0.46M &76.6  &76.6  &76.6  &76.6  &76.6 & 76.6 & ResNet-18 & 11.22M &73.2  &73.2  &73.2  & 73.2  & 73.2 \\
	\bottomrule
    \bottomrule
 
DyTox \cite{Douillard_2022_CVPR} (CVPR'2022) & ViT-Tiny & 10.71M &75.3 &73.5 &72.7 &72.3 &70.5 &72.9 & ViT-Tiny & 10.71M &77.8 &75.4 &72.9 &72.3 &74.6  \\ 
\rowcolor{lightgray}
DyTox \cite{Douillard_2022_CVPR} + \textbf{Ours}$^\ddag$& ViT-Tiny & 10.71M&\textcolor{deepred}{\textbf{75.9}} &\textcolor{deepred}{\textbf{74.0}} &\textcolor{deepred}{\textbf{72.9}} &\textcolor{deepred}{\textbf{72.6}} &\textcolor{deepred}{\textbf{70.8}} &\textcolor{deepred}{\textbf{73.2}} & ViT-Tiny & 10.71M &\textcolor{deepred}{\textbf{77.9}} &\textcolor{deepred}{\textbf{75.6}} &\textcolor{deepred}{\textbf{73.1}} &\textcolor{deepred}{\textbf{72.9}} &\textcolor{deepred}{\textbf{74.9}}  \\
\midrule
Upper Bound & ViT-Tiny & 10.71M &76.1  &76.1  &76.1  &76.1  &76.1 &76.1 & ViT-Tiny & 10.71M &79.1 &79.1 &79.1 &79.1 &79.1 \\
\bottomrule
 
\end{tabular}}
\label{tab: two_losses_cifar_imagenet100}
\vspace{-10pt}
\end{table*}

\begin{table}[t]
\centering
\setlength{\tabcolsep}{0.85mm}
\caption{Ablation studies on CIFAR-100 \cite{krizhevsky2009learning} (the top block) and ImageNet-100 \cite{5206848} (the bottom block), when we set $T=\{5, 10, 25\}$ under the settings of $\mathcal{M}=2000, \mathcal{B}= 0\%$. }
	\scalebox{0.72}{
		\begin{tabular}{l|cc|ccc|ccc}
			\toprule
\makecell[c]{Ablation Variants} & Backbone &\#Params & TSA  & GFC & GRD & 5 & 10 & 25 \\
\midrule
Baseline & ViT-Base & 85.10M & \XSolidBrush  & \XSolidBrush & \XSolidBrush &81.6 &80.2 &78.1 \\
Baseline + TSA &ViT-Base & 85.10M& \Checkmark  & \XSolidBrush & \XSolidBrush &83.1 &82.4 &81.0   \\
Baseline + TSA + GFC & ViT-Base & 85.10M& \Checkmark  & \Checkmark & \XSolidBrush &84.0 &83.7 &82.1   \\ 
\rowcolor{lightgray}
\textbf{HFC} (\textbf{Ours}) & ViT-Base & 85.10M & \Checkmark & \Checkmark & \Checkmark & \textcolor{deepred}{\textbf{86.4}} &\textcolor{deepred}{\textbf{86.3}} &\textcolor{deepred}{\textbf{85.0}}  \\

\bottomrule
\bottomrule

Baseline & ViT-Base & 85.10M& \XSolidBrush & \XSolidBrush & \XSolidBrush &82.0 &80.5 &78.8  \\
Baseline + TSA &ViT-Base & 85.10M & \Checkmark & \XSolidBrush & \XSolidBrush &83.7 &82.8 &81.2   \\
Baseline + TSA + GFC &  ViT-Base & 85.10M & \Checkmark & \XSolidBrush & \XSolidBrush &84.3 &83.4 &82.0   \\
\rowcolor{lightgray}
\textbf{HFC} (\textbf{Ours}) & ViT-Base & 85.10M & \Checkmark & \Checkmark & \Checkmark & \textcolor{deepred}{\textbf{86.1}} & \textcolor{deepred}{\textbf{85.4}} & \textcolor{deepred}{\textbf{84.0}}  \\
\bottomrule
\end{tabular}}
\label{tab: ablation_studies_cifar100}
 \vspace{-10pt}
\end{table}

\subsection{Optimization Pipeline}
Overall, in the $t$-th ($t\geq2$) learning task, the objective formulation to optimize $\Theta^t$ is expressed as follows: 
\begin{align}
	\mathcal{L}_{\mr{obj}} = \alpha_1\mathcal{L}_{\mr{FC}} + \alpha_2\mathcal{L}_{\mr{RD}},
	\label{eq: overall_objective}
\end{align}
where $\alpha_1, \alpha_2$ are hyper-parameters. $\Theta^t$ along with learnable $\mathbf{E}_0^t$ are optimized via $\mathcal{L}_{\mr{CE}}$ in Eq.~\eqref{eq: classification_loss} for the first task, and trained via $\mathcal{L}_{\mr{obj}}$ in Eq.~\eqref{eq: overall_objective} when $t\geq2$. After optimizing new model $\Theta^t$, we store $\Theta^t$ as the frozen old model $\Theta^{t-1}$ to perform the GRD loss $\mathcal{L}_{\mr{RD}}$ for the next incremental task.

\section{Experiments}\label{sec: experiments}
\subsection{Implementation Details}
For fair comparisons, we follow the same protocols utilized in baseline CIL methods \cite{Rebuffi_2017_CVPR, wu2019large, 10.1007/978-3-030-58565-5_6, 10.1007/978-3-030-58565-5_6, Ahn_2021_ICCV, Douillard_2022_CVPR, Kang_2022_CVPR} to set incremental tasks (class order), and do experiments on CIFAR-100 \cite{krizhevsky2009learning}, ImageNet-100 \cite{5206848} and ImageNet-1000 \cite{5206848}. Specifically, we consider two task settings in this paper: 1) We divide all categories of each dataset into $T$ tasks equally \cite{Rebuffi_2017_CVPR, Douillard_2022_CVPR} (\emph{i.e.}, the base class set $\mathcal{B}=0\%$); 2) We start by training our model on half of the categories, while divide the rest of categories into $T$ tasks equally \cite{10.1007/978-3-030-58565-5_6, Kang_2022_CVPR} (\emph{i.e.}, the base class set $\mathcal{B}=50\%$). When $\mathcal{B}=0\%$, we set $T=\{5, 10, 20, 25, 50\}$ for CIFAR-100 \cite{krizhevsky2009learning}, $T=\{5, 10, 20, 25\}$ for ImageNet-100 \cite{5206848} and $T=\{5, 10\}$ for ImageNet-1000 \cite{5206848}. When $\mathcal{B}=50\%$, we set $T=\{5, 10, 25, 50\}$ for CIFAR-100 \cite{krizhevsky2009learning} and ImageNet-100 \cite{5206848}.

As introduced in \cite{Rebuffi_2017_CVPR, wu2019large}, the size of memory $\mathcal{M}$ for all comparison methods is fixed as 2,000 for CIFAR-100 \cite{krizhevsky2009learning} and ImageNet-100 \cite{5206848}, and 20,000 for ImageNet-1000 \cite{5206848}. We follow iCaRL \cite{Rebuffi_2017_CVPR} to update memory $\mathcal{M}$ and use global memory for \cite{Douillard_2022_CVPR}. Besides, we employ the same data augmentation proposed in DyTox \cite{Douillard_2022_CVPR} for all comparison methods, but don't use mixup technology. Following \cite{10.1007/978-3-030-58565-5_6, Kang_2022_CVPR}, we also consider another memory setting (\emph{i.e.}, storing 20 exemplars per class), and set $T=\{5, 10\}, \mathcal{B}=50\%$ for comparisons on CIFAR-100 \cite{krizhevsky2009learning} (see Tab.~\ref{tab: comparison_exp_cifar100_exemplar20perclass}). 
For network architecture, we use ViT-Base \cite{dosovitskiy2020vit} as feature extractor containing $l_m=11$ MSA blocks and $l_t=1$ TSA block, where the parameters are initialized via \cite{He_2022_CVPR}. The task classifier includes only one fully-connected layer. The SGD optimizer is employed to optimize our model, where the learning rate is initialized as $6.25\times 10^{-5}$. We apply two plug-and-play losses (\emph{i.e.}, $\mathcal{L}_{\mr{FC}}$ and $\mathcal{L}_{\mr{RD}}$) to existing CIL methods \cite{Rebuffi_2017_CVPR, Douillard_2022_CVPR} using ResNet-18 \cite{7780459}, ResNet-32 \cite{7780459} or ViT-Tiny \cite{dosovitskiy2020vit} as backbone. For evaluation, top-1 accuracy is used to compare performances of our model and baseline methods.

\subsection{Comparison Experiments}

\textbf{Comparison Results:} Tabs.~\ref{tab: comparison_exp_cifar100}--\ref{tab: comparison_exp_image1000} present comparison results of our HFC model and off-the-shelf CIL methods on CIFAR-100 \cite{krizhevsky2009learning}, ImageNet-100 \cite{5206848} and ImageNet-1000 \cite{5206848}. Our model significantly outperforms existing CIL methods \cite{Rebuffi_2017_CVPR, wu2019large, 10.1007/978-3-030-58565-5_6, Ahn_2021_ICCV, Douillard_2022_CVPR, Kang_2022_CVPR, 10.1007/978-3-031-19806-9_23} by $0.7\%\sim16.6\%$ accuracy under various incremental settings, when we use the same backbone (\emph{i.e.}, ViT-Base \cite{dosovitskiy2020vit}) as feature extractor for fair comparisons. Such large performance improvement illustrates the superiority of our model to address forgetting heterogeneity among easy-to-forget and hard-to-forget old categories from both representation and gradient aspects.

\textbf{Improvement of Plug-and-Play Losses:}
Tab.~\ref{tab: two_losses_cifar_imagenet100} shows large performance improvement of existing knowledge-based CIL methods \cite{Rebuffi_2017_CVPR, wu2019large, 10.1007/978-3-030-58565-5_6, 10.1007/978-3-031-19812-0_7, Douillard_2022_CVPR, Kang_2022_CVPR, 10.1007/978-3-031-19806-9_23}, when we apply the proposed plug-and-play losses (\emph{i.e.}, $\mathcal{L}_{\mr{FC}}$ and $\mathcal{L}_{\mr{RD}}$) to them. The proposed losses $\mathcal{L}_{\mr{FC}}$ and $\mathcal{L}_{\mr{RD}}$ help existing CIL methods to address heterogeneous forgetting from gradient aspect, thus largely improving their performance. Besides, it validates generalization and robustness of our HFC model.

\begin{figure*}[t]
\centering
\includegraphics[trim = 16mm 72mm 18mm 72mm, clip, width=495pt, height=120pt]
{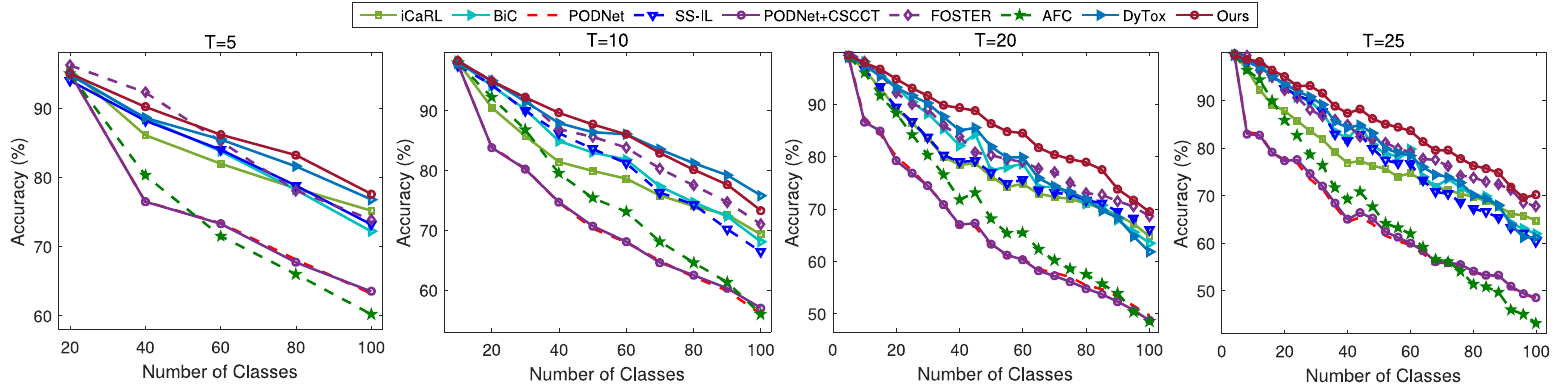}
\vspace{-15pt}
\caption{Analysis of task-wise performance comparisons on CIFAR-100 \cite{krizhevsky2009learning}. We set $\mathcal{M}=2000, \mathcal{B}=0\%$ when the backbone is ViT-Base. }
\label{fig: incremental_tasks_CIFAR100}
\vspace{-10pt}
\end{figure*}

\begin{figure*}[t]
\centering
\includegraphics[trim = 12mm 68mm 14mm 68mm, clip, width=495pt, height=120pt]
{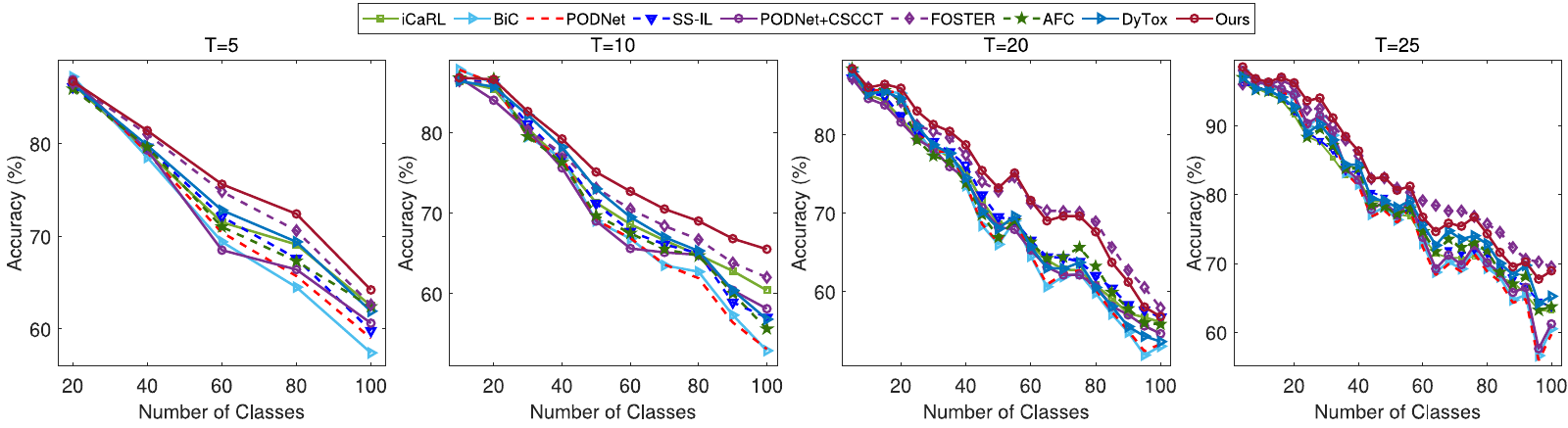}
\vspace{-15pt}
\caption{Analysis of task-wise performance on ImageNet-100 \cite{5206848}. We set $\mathcal{M}=2000, \mathcal{B}=0\%$ when the backbone is ViT-Base. }
\label{fig: incremental_tasks_Imagenet_subnet}
\vspace{-10pt}
\end{figure*}

\subsection{Ablation Studies}
As shown in Tab.~\ref{tab: ablation_studies_cifar100}, we present ablation studies on CIFAR-100 \cite{krizhevsky2009learning} and ImageNet-100 \cite{5206848}
to investigate the effectiveness of each module. TSA, GFC and GRD are task-semantic aggregation blocks, the gradient-balanced forgetting compensation loss $\mathcal{L}_{\mr{FC}}$ and gradient-balanced relation distillation loss $\mathcal{L}_{\mr{RD}}$. Baseline denotes the performance of our model using the traditional classification loss and knowledge distillation loss proposed in DyTox (CVPR'2022) \cite{Douillard_2022_CVPR}, but without the TSA, GFC and GRD modules. Compared with Ours, the performance of all ablation variants degrades significantly. It validates superiority of each module to overcome forgetting heterogeneity of different old categories.

\begin{table}[t]
\centering
\setlength{\tabcolsep}{1.0mm}
\caption{Results on CIFAR-100 \cite{krizhevsky2009learning} when storing 20 exemplars for each class. We set $T=\{5, 10\}$ for $ \mathcal{B}=50\%$, and \textbf{Ours$^\ddag$} denotes the proposed plug-and-play losses $\mathcal{L}_{\rm{FC}}$ and $\mathcal{L}_{\rm{RD}}$. }
\scalebox{0.64}{
\begin{tabular}{l|cc|cc|cl}
\toprule
  \makecell[c]{Comparison Methods}& Backbone & \#Params & 5 & 10 & Avg. & Imp. \\
\midrule
iCaRL \cite{Rebuffi_2017_CVPR} (CVPR'2017) & ViT-Base & 85.10M &78.8 &76.2 &77.5  &$\Uparrow$1.9 \\
BiC \cite{wu2019large} (CVPR'2019) & ViT-Base & 85.10M &75.3 &74.5 &74.9  &$\Uparrow$4.5\\
PODNet \cite{10.1007/978-3-030-58565-5_6} (ECCV'2020) & ViT-Base & 85.10M &71.6 &71.6 &71.6  &$\Uparrow$7.8 \\
SS-IL \cite{Ahn_2021_ICCV} (ICCV'2021) & ViT-Base& 85.10M &73.0 &71.8 &72.4  &$\Uparrow$7.0  \\
PODNet \cite{10.1007/978-3-030-58565-5_6} + CSCCT \cite{10.1007/978-3-031-19812-0_7} (ECCV'2022) & ViT-Base & 85.10M &71.5 &71.7 &71.6 &$\Uparrow$7.8\\
FOSTER \cite{10.1007/978-3-031-19806-9_23} (ECCV'2022) & ViT-Base & 85.10M &78.8 &78.7 &78.8 &$\Uparrow$0.6\\
    AFC \cite{Kang_2022_CVPR} (CVPR'2022) & ViT-Base & 85.10M &68.2 &65.3 &66.8 &$\Uparrow$12.6  \\
DyTox \cite{Douillard_2022_CVPR} (CVPR'2022) & ViT-Base & 85.10M &79.3 &78.4 &78.8 &$\Uparrow$0.6 \\ 
    \midrule
    \rowcolor{lightgray}
    \textbf{HFC} (\textbf{Ours}) & ViT-Base & 85.10M & \textcolor{deepred}{\textbf{79.8}} & \textcolor{deepred}{\textbf{78.9}} & \textcolor{deepred}{\textbf{79.4}}  &\textbf{$\mathrm{-}$}  \\
    \midrule
    Upper Bound & ViT-Base & 85.10M &94.2 &94.2 &94.2 &\textbf{$\mathrm{-}$} \\
\bottomrule
\bottomrule
iCaRL \cite{Rebuffi_2017_CVPR} (CVPR'2017) & ResNet-32 & 0.46M &56.6 &53.2 &54.9 &$\Uparrow$0.7 \\
        \rowcolor{lightgray}
        iCaRL \cite{Rebuffi_2017_CVPR} + \textbf{Ours}$^\ddag$& ResNet-32 & 0.46M & \textcolor{deepred}{\textbf{57.3}} & \textcolor{deepred}{\textbf{53.9}} & \textcolor{deepred}{\textbf{55.6}} &\textbf{$\mathrm{-}$} \\
 \midrule
  PODNet \cite{10.1007/978-3-030-58565-5_6} + CSCCT \cite{10.1007/978-3-031-19812-0_7} (ECCV'2022) & ResNet-32 & 0.46M  & 62.2 &61.1 &61.7 &$\Uparrow$0.6\\
   \rowcolor{lightgray}
   PODNet \cite{10.1007/978-3-030-58565-5_6} + CSCCT \cite{10.1007/978-3-031-19812-0_7} + \textbf{Ours}$^\ddag$ & ResNet-32 & 0.46M & \textcolor{deepred}{\textbf{62.9}} & \textcolor{deepred} {\textbf{61.6}}  & \textcolor{deepred}{\textbf{62.3}} &\textbf{$\mathrm{-}$}  \\
   \midrule
 FOSTER \cite{10.1007/978-3-031-19806-9_23} (ECCV'2022) & ResNet-32 & 0.46M &67.3 &66.4 &66.8 &$\Uparrow$0.5 \\
 \rowcolor{lightgray}
   FOSTER \cite{10.1007/978-3-031-19806-9_23} + \textbf{Ours}$^\ddag$ & ResNet-32 & 0.46M & \textcolor{deepred}{\textbf{67.7}}  & \textcolor{deepred}{\textbf{66.6}}  & \textcolor{deepred}{\textbf{67.2}} &\textbf{$\mathrm{-}$}   \\
        \midrule
        AFC \cite{Kang_2022_CVPR} (CVPR'2022) &ResNet-32 & 0.46M &63.8 &63.6 &63.7 &$\Uparrow$0.7  \\
        \rowcolor{lightgray}
   AFC \cite{Kang_2022_CVPR} + \textbf{Ours}$^\ddag$ & ResNet-32 & 0.46M & \textcolor{deepred}{\textbf{64.4}} & \textcolor{deepred}{\textbf{64.3}}  & \textcolor{deepred}{\textbf{64.4}}&\textbf{$\mathrm{-}$} \\
   \midrule
   Upper Bound & ResNet-32 & 0.46M &76.6 &76.6 &76.6&\textbf{$\mathrm{-}$}   \\
    \bottomrule
 \bottomrule
   DyTox \cite{Douillard_2022_CVPR} (CVPR'2022) & ViT-Tiny & 10.71M &67.9  &66.2   &67.1  &$\Uparrow$0.5  \\ 
\rowcolor{lightgray}
DyTox \cite{Douillard_2022_CVPR} + \textbf{Ours}$^\ddag$& ViT-Tiny & 10.71M & \textcolor{deepred}{\textbf{68.5}} & \textcolor{deepred} {\textbf{66.7}}  & \textcolor{deepred}{\textbf{67.6}}&\textbf{$\mathrm{-}$}\\
\midrule
Upper Bound & ViT-Tiny & 10.71M &76.1 &76.1 &76.1 &\textbf{$\mathrm{-}$} \\
\bottomrule
\end{tabular}}
\label{tab: comparison_exp_cifar100_exemplar20perclass}
\vspace{-10pt}
\end{table}

\subsection{Analysis of Task-Wise Comparisons}
Figs.~\ref{fig: incremental_tasks_CIFAR100}--\ref{fig: incremental_tasks_Imagenet_subnet} present task-wise performance comparisons between our model and other CIL methods \cite{Rebuffi_2017_CVPR, wu2019large, 10.1007/978-3-030-58565-5_6, Douillard_2022_CVPR, Kang_2022_CVPR} when we set backbone as ViT-Base and different tasks as $T=\{5, 10, 20, 25\}$. Our HFC model outperforms baseline methods for most task-wise comparisons, which illustrates the superior performance of our model to surmount heterogeneous forgetting from representation and gradient aspects. To identify new classes continually, the task-semantic aggregation (TSA) block explores task-shared global representations to alleviate forgetting from representation aspect, while $\mathcal{L}_{\mr{FC}}$ and $\mathcal{L}_{\mr{RD}}$ can achieve gradient-balanced compensation.

\subsection{Analysis of Forgetting Heterogeneity}
Tab.~\ref{tab: forgetting_heterogeneity} presents forgetting heterogeneity (FH) of different old classes via measuring their variance of gradient updating. We formulate forgetting heterogeneity (FH) as $\mr{FH} = \frac{1}{T} \sum \nolimits_{t=1}^{T} \big(\frac{1}{S_t}\sum\nolimits_{i=1}^{S_t}(|\Gamma_{i}^t| - \sum\nolimits_{\eta=1}^t \Gamma_\eta\cdot \mathbb{I}_{\mf{y}_{i}^t\in\mathcal{Y}^{\eta}})^2$ \big), where $\Gamma_{i}^t$ and $\Gamma_\eta$ are obtained from Eqs.~\eqref{eq: gradient_value}--\eqref{eq: task_specific_gradient}, and $S_t$ is number of samples in test set. Tab.~\ref{tab: forgetting_heterogeneity} verifies superiority of our model against other CIL methods to tackle heterogeneous forgetting. It also shows effectiveness of all modules to collaboratively minimize forgetting heterogeneity.

\begin{table}[t]
\centering
\setlength{\tabcolsep}{1.1mm}
\caption{Analysis of forgetting heterogeneity (FH) on CIFAR-100 \cite{krizhevsky2009learning} when $T=5, \mathcal{B}=0\%$. AC denotes the averaged accuracy.} 
	\scalebox{0.75}{
		\begin{tabular}{c|c|ccc|cc}
			\toprule
$\mathcal{M}=2000, \mathcal{B}=0\%$ & Backbone & TSA  & GFC & GRD & FH & AC \\
\midrule
iCaRL \cite{Rebuffi_2017_CVPR} (CVPR'2017) & ViT-Base & -- & -- & -- &131.5  & 83.5 \\
FOSTER \cite{10.1007/978-3-031-19806-9_23} (ECCV'2022)  &ViT-Base & -- & -- & -- &119.4 & 85.1 \\
AFC \cite{Kang_2022_CVPR} (CVPR'2022) & ViT-Base & -- & -- & -- &108.5  & 74.6 \\
DyTox \cite{Douillard_2022_CVPR} (CVPR’2022) &ViT-Base & -- & -- & -- &105.1 & 85.5  \\
\midrule
 Baseline & ViT-Base & \XSolidBrush  & \XSolidBrush & \XSolidBrush &128.9 &81.6 \\
Baseline + TSA &ViT-Base & \Checkmark  & \XSolidBrush & \XSolidBrush &125.4 &83.1  \\
Baseline + TSA + GFC & ViT-Base & \Checkmark  & \Checkmark & \XSolidBrush &120.5 &84.0  \\
\textbf{HFC} (\textbf{Ours}) & ViT-Base & \Checkmark & \Checkmark & \Checkmark  &\textcolor{deepred}{\textbf{103.6}} & \textcolor{deepred}{\textbf{86.4}}   \\
\bottomrule
\end{tabular}}
\label{tab: forgetting_heterogeneity}
 \vspace{-12pt}
\end{table}

\section{Conclusion}\label{sec: conclusion}
In this paper, we develop a novel Heterogeneous Forgetting Compensation (HFC) model to surmount heterogeneous forgetting from representation and gradient aspects. To be specific, a task-semantic aggregation block is designed to tackle heterogeneous forgetting from representation aspect via exploring task-shared global representations. Moreover, we propose a gradient-balanced forgetting compensation loss and a relation distillation loss to compensate forgetting heterogeneity from gradient aspect via performing balanced gradient propagation and distilling heterogeneous class relations. Experiments verify the superiority of our proposed HFC model against baseline methods. We will further consider addressing noisy forgetting heterogeneity in the future.

{\small
\bibliographystyle{ieee_fullname}
\bibliography{HFC}

\begin{thebibliography}{10}\itemsep=-1pt

\bibitem{9156310}
Davide Abati, Jakub Tomczak, Tijmen Blankevoort, Simone Calderara, Rita
  Cucchiara, and Babak~Ehteshami Bejnordi.
\newblock Conditional channel gated networks for task-aware continual learning.
\newblock In {\em CVPR}, pages 3930--3939, 2020.

\bibitem{Ahn_2021_ICCV}
Hongjoon Ahn, Jihwan Kwak, Subin Lim, Hyeonsu Bang, Hyojun Kim, and Taesup
  Moon.
\newblock Ss-il: Separated softmax for incremental learning.
\newblock In {\em ICCV}, pages 844--853, October 2021.

\bibitem{10.1007/978-3-030-01219-9_9}
Rahaf Aljundi, Francesca Babiloni, Mohamed Elhoseiny, Marcus Rohrbach, and
  Tinne Tuytelaars.
\newblock Memory aware synapses: Learning what (not) to forget.
\newblock In Vittorio Ferrari, Martial Hebert, Cristian Sminchisescu, and Yair
  Weiss, editors, {\em ECCV}, pages 144--161, 2018.

\bibitem{10.1007/978-3-031-19812-0_7}
Arjun Ashok, K.~J. Joseph, and Vineeth~N. Balasubramanian.
\newblock Class-incremental learning with cross-space clustering and controlled
  transfer.
\newblock In {\em ECCV}, pages 105--122, 2022.

\bibitem{9891836}
Matteo Boschini, Lorenzo Bonicelli, Pietro Buzzega, Angelo Porrello, and Simone
  Calderara.
\newblock Class-incremental continual learning into the extended der-verse.
\newblock {\em IEEE Transactions on Pattern Analysis and Machine Intelligence},
  pages 1--16, 2022.

\bibitem{10.1007/978-3-030-01258-8_15}
Francisco~M. Castro, Manuel~J. Mar{\'i}n-Jim{\'e}nez, Nicol{\'a}s Guil,
  Cordelia Schmid, and Karteek Alahari.
\newblock End-to-end incremental learning.
\newblock In Vittorio Ferrari, Martial Hebert, Cristian Sminchisescu, and Yair
  Weiss, editors, {\em ECCV}, pages 241--257, Cham, 2018. Springer
  International Publishing.

\bibitem{chen2023vlp}
Fei-Long Chen, Du-Zhen Zhang, Ming-Lun Han, Xiu-Yi Chen, Jing Shi, Shuang Xu,
  and Bo Xu.
\newblock Vlp: A survey on vision-language pre-training.
\newblock {\em Machine Intelligence Research}, 20(1):38--56, 2023.

\bibitem{9815145}
Huitong Chen, Yu Wang, and Qinghua Hu.
\newblock Multi-granularity regularized re-balancing for class incremental
  learning.
\newblock {\em IEEE Transactions on Knowledge and Data Engineering}, pages
  1--15, 2022.

\bibitem{d2021convit}
St{\'e}phane d'Ascoli, Hugo Touvron, Matthew Leavitt, Ari Morcos, Giulio
  Biroli, and Levent Sagun.
\newblock Convit: Improving vision transformers with soft convolutional
  inductive biases.
\newblock In {\em ICML}, 2021.

\bibitem{5206848}
Jia Deng, Wei Dong, Richard Socher, Li-Jia Li, Kai Li, and Li Fei-Fei.
\newblock Imagenet: A large-scale hierarchical image database.
\newblock In {\em CVPR}, pages 248--255, 2009.

\bibitem{9616392_Dong}
Jiahua Dong, Yang Cong, Gan Sun, Zhen Fang, and Zhengming Ding.
\newblock Where and how to transfer: Knowledge aggregation-induced
  transferability perception for unsupervised domain adaptation.
\newblock {\em IEEE Transactions on Pattern Analysis and Machine Intelligence},
  pages 11--20, 2021.

\bibitem{Dong_2022_CVPR}
Jiahua Dong, Lixu Wang, Zhen Fang, Gan Sun, Shichao Xu, Xiao Wang, and Qi Zhu.
\newblock Federated class-incremental learning.
\newblock In {\em CVPR}, pages 10164--10173, June 2022.

\bibitem{dong2023federated_FISS}
Jiahua Dong, Duzhen Zhang, Yang Cong, Wei Cong, Henghui Ding, and Dengxin Dai.
\newblock Federated incremental semantic segmentation.
\newblock In {\em CVPR}, pages 3934--3943, June 2023.

\bibitem{DongXin_2022_CVPR}
Xin Dong, Junfeng Guo, Ang Li, Wei-Te Ting, Cong Liu, and H.T. Kung.
\newblock Neural mean discrepancy for efficient out-of-distribution detection.
\newblock In {\em CVPR}, pages 19217--19227, June 2022.

\bibitem{dosovitskiy2020vit}
Alexey Dosovitskiy, Lucas Beyer, Alexander Kolesnikov, Dirk Weissenborn,
  Xiaohua Zhai, Thomas Unterthiner, Mostafa Dehghani, Matthias Minderer, Georg
  Heigold, Sylvain Gelly, Jakob Uszkoreit, and Neil Houlsby.
\newblock An image is worth 16x16 words: Transformers for image recognition at
  scale.
\newblock {\em ICLR}, 2021.

\bibitem{douillard2021plop}
Arthur Douillard, Yifu Chen, Arnaud Dapogny, and Matthieu Cord.
\newblock Plop: Learning without forgetting for continual semantic
  segmentation.
\newblock In {\em CVPR}, 2021.

\bibitem{10.1007/978-3-030-58565-5_6}
Arthur Douillard, Matthieu Cord, Charles Ollion, Thomas Robert, and Eduardo
  Valle.
\newblock Podnet: Pooled outputs distillation for small-tasks incremental
  learning.
\newblock In {\em ECCV}, pages 86--102, 2020.

\bibitem{Douillard_2022_CVPR}
Arthur Douillard, Alexandre Ram\'e, Guillaume Couairon, and Matthieu Cord.
\newblock Dytox: Transformers for continual learning with dynamic token
  expansion.
\newblock In {\em CVPR}, pages 9285--9295, June 2022.

\bibitem{NEURIPS2022_f0e91b13}
Zhen Fang, Yixuan Li, Jie Lu, Jiahua Dong, Bo Han, and Feng Liu.
\newblock Is out-of-distribution detection learnable?
\newblock In {\em NeurIPS}, volume~35, pages 37199--37213, 2022.

\bibitem{10.5555/3305381.3305510}
Jonas Gehring, Michael Auli, David Grangier, Denis Yarats, and Yann~N. Dauphin.
\newblock Convolutional sequence to sequence learning.
\newblock In {\em ICML}, page 1243–1252, 2017.

\bibitem{NIPS2014_5ca3e9b1}
Ian Goodfellow, Jean Pouget-Abadie, Mehdi Mirza, Bing Xu, David Warde-Farley,
  Sherjil Ozair, Aaron Courville, and Yoshua Bengio.
\newblock Generative adversarial nets.
\newblock In {\em NeurIPS}, volume~27, 2014.

\bibitem{He_2022_CVPR}
Kaiming He, Xinlei Chen, Saining Xie, Yanghao Li, Piotr Doll\'ar, and Ross
  Girshick.
\newblock Masked autoencoders are scalable vision learners.
\newblock In {\em CVPR}, pages 16000--16009, June 2022.

\bibitem{7780459}
Kaiming He, Xiangyu Zhang, Shaoqing Ren, and Jian Sun.
\newblock Deep residual learning for image recognition.
\newblock In {\em CVPR}, pages 770--778, 2016.

\bibitem{44873_Distilling}
Geoffrey Hinton, Oriol Vinyals, and Jeffrey Dean.
\newblock Distilling the knowledge in a neural network.
\newblock In {\em NeurIPS Workshop}, 2015.

\bibitem{Hou_2019_CVPR}
Saihui Hou, Xinyu Pan, Chen~Change Loy, Zilei Wang, and Dahua Lin.
\newblock Learning a unified classifier incrementally via rebalancing.
\newblock In {\em CVPR}, June 2019.

\bibitem{6482546}
Weiming Hu, Xi Li, Guodong Tian, Stephen Maybank, and Zhongfei Zhang.
\newblock An incremental dpmm-based method for trajectory clustering, modeling,
  and retrieval.
\newblock {\em IEEE Transactions on Pattern Analysis and Machine Intelligence},
  35(5):1051--1065, 2013.

\bibitem{Hu_2021_CVPR}
Xinting Hu, Kaihua Tang, Chunyan Miao, Xian-Sheng Hua, and Hanwang Zhang.
\newblock Distilling causal effect of data in class-incremental learning.
\newblock In {\em CVPR}, 2021.

\bibitem{10.5555/3454287.3455512}
Ching-Yi Hung, Cheng-Hao Tu, Cheng-En Wu, Chien-Hung Chen, Yi-Ming Chan, and
  Chu-Song Chen.
\newblock Compacting, picking and growing for unforgetting continual learning.
\newblock In {\em Advances in Neural Information Processing Systems}, pages
  13647--13657, 2019.

\bibitem{10.1007/978-3-030-58517-4_41}
Ahmet Iscen, Jeffrey Zhang, Svetlana Lazebnik, and Cordelia Schmid.
\newblock Memory-efficient incremental learning through feature adaptation.
\newblock In Andrea Vedaldi, Horst Bischof, Thomas Brox, and Jan-Michael Frahm,
  editors, {\em ECCV}, pages 699--715, 2020.

\bibitem{jaegle2021perceiver}
Andrew Jaegle, Sebastian Borgeaud, Jean-Baptiste Alayrac, Carl Doersch, Catalin
  Ionescu, and et al.
\newblock Perceiver io: A general architecture for structured inputs \&
  outputs.
\newblock In {\em ICLR}, 2022.

\bibitem{joseph2022Energy}
KJ Joseph, Salman Khan, Fahad~Shahbaz Khan, Rao~Muhammad Anwar, and Vineeth
  Balasubramanian.
\newblock Energy-based latent aligner for incremental learning.
\newblock In {\em CVPR}, 2022.

\bibitem{Kang_2022_CVPR}
Minsoo Kang, Jaeyoo Park, and Bohyung Han.
\newblock Class-incremental learning by knowledge distillation with adaptive
  feature consolidation.
\newblock In {\em CVPR}, pages 16071--16080, June 2022.

\bibitem{krizhevsky2009learning}
Alex Krizhevsky.
\newblock Learning multiple layers of features from tiny images.
\newblock {\em Technical report}, pages 32--33, 2009.

\bibitem{8107520}
Zhizhong Li and Derek Hoiem.
\newblock Learning without forgetting.
\newblock {\em IEEE Transactions on Pattern Analysis and Machine Intelligence},
  40(12):2935--2947, 2018.

\bibitem{liu2022deja}
Chenxi Liu, Lixu Wang, Lingjuan Lyu, Chen Sun, Xiao Wang, and Qi Zhu.
\newblock Deja vu: Continual model generalization for unseen domains.
\newblock In {\em ICLR}, 2022.

\bibitem{Liu2023Online}
Yaoyao Liu, Yingying Li, Bernt Schiele, and Qianru Sun.
\newblock Online hyperparameter optimization for class-incremental learning.
\newblock {\em AAAI}, 37(7):8906--8913, Jun. 2023.

\bibitem{DBLP:conf/nips/LiuSS21}
Yaoyao Liu, Bernt Schiele, and Qianru Sun.
\newblock {RMM:} reinforced memory management for class-incremental learning.
\newblock In {\em NeurIPS}, pages 3478--3490, 2021.

\bibitem{8569992}
John~M. Pierre.
\newblock Incremental lifelong deep learning for autonomous vehicles.
\newblock In {\em ITSC}, pages 3949--3954, 2018.

\bibitem{NEURIPS2019_83da7c53}
Jathushan Rajasegaran, Munawar Hayat, Salman~H Khan, Fahad~Shahbaz Khan, and
  Ling Shao.
\newblock Random path selection for continual learning.
\newblock In {\em NeurIPS}, 2019.

\bibitem{Rebuffi_2017_CVPR}
Sylvestre-Alvise Rebuffi, Alexander Kolesnikov, Georg Sperl, and Christoph~H.
  Lampert.
\newblock icarl: Incremental classifier and representation learning.
\newblock In {\em CVPR}, 2017.

\bibitem{pmlr-v80-serra18a}
Joan Serra, Didac Suris, Marius Miron, and Alexandros Karatzoglou.
\newblock Overcoming catastrophic forgetting with hard attention to the task.
\newblock In {\em ICML}, volume~80, pages 4548--4557, 2018.

\bibitem{tang2022learning}
Yu-Ming Tang, Yi-Xing Peng, and Wei-Shi Zheng.
\newblock Learning to imagine: Diversify memory for incremental learning using
  unlabeled data.
\newblock In {\em CVPR}, 2022.

\bibitem{tang2022virtual}
Zhenheng Tang, Yonggang Zhang, Shaohuai Shi, Xin He, Bo Han, and Xiaowen Chu.
\newblock Virtual homogeneity learning: Defending against data heterogeneity in
  federated learning.
\newblock In {\em ICML}, pages 21111--21132, 2022.

\bibitem{10.1007/978-3-030-58529-7_16}
Xiaoyu Tao, Xinyuan Chang, Xiaopeng Hong, Xing Wei, and Yihong Gong.
\newblock Topology-preserving class-incremental learning.
\newblock In Andrea Vedaldi, Horst Bischof, Thomas Brox, and Jan-Michael Frahm,
  editors, {\em ECCV}, pages 254--270, Cham, 2020. Springer International
  Publishing.

\bibitem{Tiwari_2022_CVPR}
Rishabh Tiwari, Krishnateja Killamsetty, Rishabh Iyer, and Pradeep Shenoy.
\newblock Gcr: Gradient coreset based replay buffer selection for continual
  learning.
\newblock In {\em CVPR}, pages 99--108, June 2022.

\bibitem{9710634}
Hugo Touvron, Matthieu Cord, Alexandre Sablayrolles, Gabriel Synnaeve, and
  Hervé Jégou.
\newblock Going deeper with image transformers.
\newblock In {\em ICCV}, pages 32--42, 2021.

\bibitem{10.1007/978-3-031-19806-9_23}
Fu-Yun Wang, Da-Wei Zhou, Han-Jia Ye, and De-Chuan Zhan.
\newblock Foster: Feature boosting and compression for class-incremental
  learning.
\newblock In {\em ECCV}, pages 398--414, 2022.

\bibitem{wang2021addressing}
Lixu Wang, Shichao Xu, Xiao Wang, and Qi Zhu.
\newblock Addressing class imbalance in federated learning.
\newblock In {\em AAAI}, volume~35, pages 10165--10173, 2021.

\bibitem{Wang_ICLR2022}
Liyuan Wang, Xingxing Zhang, Kuo Yang, Longhui Yu, Chongxuan Li, Lanqing Hong,
  Shifeng Zhang, Zhenguo Li, Yi Zhong, and Jun Zhu.
\newblock Memory replay with data compression for continual learning.
\newblock In {\em ICLR}, 2022.

\bibitem{wu2019large}
Yue Wu, Yinpeng Chen, Lijuan Wang, Yuancheng Ye, Zicheng Liu, Yandong Guo, and
  Yun Fu.
\newblock Large scale incremental learning.
\newblock In {\em CVPR}, pages 374--382, 2019.

\bibitem{9578918}
Ziyang Wu, Christina Baek, Chong You, and Yi Ma.
\newblock Incremental learning via rate reduction.
\newblock In {\em CVPR}, pages 1125--1133, 2021.

\bibitem{xu2023wave}
Rongtao Xu, Changwei Wang, Shibiao Xu, Weiliang Meng, and Xiaopeng Zhang.
\newblock Wave-like class activation map with representation fusion for
  weakly-supervised semantic segmentation.
\newblock {\em IEEE Transactions on Multimedia}, 2023.

\bibitem{xu2023rssformer}
Rongtao Xu, Changwei Wang, Jiguang Zhang, Shibiao Xu, Weiliang Meng, and
  Xiaopeng Zhang.
\newblock Rssformer: Foreground saliency enhancement for remote sensing
  land-cover segmentation.
\newblock {\em IEEE Transactions on Image Processing}, 32:1052--1064, 2023.

\bibitem{Yan_2021_CVPR}
Shipeng Yan, Jiangwei Xie, and Xuming He.
\newblock Der: Dynamically expandable representation for class incremental
  learning.
\newblock In {\em CVPR}, pages 3014--3023, 2021.

\bibitem{yoon2018lifelong}
Jaehong Yoon, Eunho Yang, Jeongtae Lee, and Sung~Ju Hwang.
\newblock Lifelong learning with dynamically expandable networks.
\newblock In {\em ICLR}, 2018.

\bibitem{10.5555/3305890.3306093}
Friedemann Zenke, Ben Poole, and Surya Ganguli.
\newblock Continual learning through synaptic intelligence.
\newblock In {\em ICML}, page 3987–3995, 2017.

\bibitem{zhang2023dualgats}
Duzhen Zhang, Feilong Chen, and Xiuyi Chen.
\newblock Dualgats: Dual graph attention networks for emotion recognition in
  conversations.
\newblock In {\em ACL}, pages 7395--7408, 2023.

\bibitem{zhang2021causaladv}
Yonggang Zhang, Mingming Gong, Tongliang Liu, Gang Niu, Xinmei Tian, Bo Han,
  Bernhard Sch{\"o}lkopf, and Kun Zhang.
\newblock Causaladv: Adversarial robustness through the lens of causality.
\newblock {\em ICLR}, 2022.

\bibitem{9156766}
Bowen Zhao, Xi Xiao, Guojun Gan, Bin Zhang, and Shu-Tao Xia.
\newblock Maintaining discrimination and fairness in class incremental
  learning.
\newblock In {\em CVPR}, pages 13205--13214, 2020.

\bibitem{zhao2021mgsvf}
Hanbin Zhao, Yongjian Fu, Mintong Kang, Qi Tian, Fei Wu, and Xi Li.
\newblock Mgsvf: Multi-grained slow vs. fast framework for few-shot
  class-incremental learning.
\newblock {\em IEEE Transactions on Pattern Analysis and Machine Intelligence},
  2021.

\bibitem{zhao2022rbc}
Hanbin Zhao, Fengyu Yang, Xinghe Fu, and Xi Li.
\newblock Rbc: Rectifying the biased context in continual semantic
  segmentation.
\newblock In {\em ECCV}, pages 55--72, 2022.

\end{thebibliography}
}

\newpage
\appendix
\section{Supplementary Material}
This supplementary material presents detailed optimization pipeline of the proposed HFC model in Section~\ref{sec: optimization_pipeline}, analysis of incremental tasks in Section~\ref{sec: experimental_results}, and stable convergence analysis in Section~\ref{sec: convergence}. The code is available at \url{https://github.com/JiahuaDong/HFC}.

\subsection{Optimization Pipeline of Our HFC Model}
\label{sec: optimization_pipeline}

The optimization of our HFC to learn new classes consecutively is summarized in \textbf{Algorithm} \ref{alg: optimization}. $\Theta^t$ along with learnable $\mathbf{E}_0^t$ are optimized via $\mathcal{L}_{\mr{CE}}$ in Eq.~\eqref{eq: classification_loss} for the first task, and trained via $\mathcal{L}_{\mr{obj}}$ in Eq.~\eqref{eq: overall_objective} when $t\geq2$. After optimizing $\Theta^t$ in the $t$-th task, we store $\Theta^t$ as the frozen old model $\Theta^{t-1}$ to perform the gradient-balanced relation distillation loss $\mathcal{L}_{\mr{RD}}$ in Eq.~\eqref{eq: class_relation_distillation} for the next learning task. Meanwhile, the exemplar memory $\mathcal{M}$ is updated via following iCaRL \cite{Rebuffi_2017_CVPR} to replay only few samples (\emph{i.e.}, $\frac{|\mathcal{M}|}{K^o+K^t}$) of each learned class for the next task, or following PODNet \cite{10.1007/978-3-030-58565-5_6} to store 20 exemplars for each learned class.

\subsection{Task-wise Performance Comparison}\label{sec: experimental_results}

In order to comprehensively evaluate the effectiveness of our proposed HFC model, top-1 accuracy is employed to compare our model with other state-of-the-art CIL methods. As shown in Tabs.~\ref{tab: comparison_exp_img100}--\ref{tab: comparison_exp_img1000}, our model achieves significant improvement over other methods by $1.4\%\sim13.2\%$ in terms of top-1 average accuracy. Specially, compared to the previous methods, the performance gap increases as the number of stages grows. It reflects that our HFC model is more effective to deal with challenging tasks and achieve solid improvement for all incremental tasks.  In addition, Our HFC model achieves superior performance compared to the baseline methods for most of the tasks, which shows that our method can effectively address the catastrophic forgetting of old classes from both representation and gradient aspects.

As shown in Tab.~\ref{tab: two_losses_cifar}, we present comparison experiments in terms of top-1 accuracy between our HFC model and other CIL baselines on CIFAR-100 \cite{krizhevsky2009learning},  when the number of tasks is 10. We apply two plug-and-play losses  (\emph{i.e.}, $\mathcal{L}_{\mr{FC}}$ and $\mathcal{L}_{\mr{RD}}$) to existing CIL methods. The experimental results show that the proposed plug-and-play losses help current CIL methods to overcome heterogeneous forgetting from a gradient perspective. More importantly, as shown in Tab.~\ref{tab: two_losses_cifar}, our proposed losses help the existing state-of-the-art CIL methods significantly improve by $0.5\%\sim6.7\%$ in terms of top-1 averaged accuracy on  CIFAR-100 \cite{krizhevsky2009learning}. It verifies stable generalization of our HFC model to address heterogeneous catastrophic forgetting.

\renewcommand{\algorithmicrequire}{\textbf{Input:}}
\renewcommand{\algorithmicensure}{\textbf{Output:}}
\begin{algorithm}[t]			
	\caption{Optimization pipeline of HFC.} 
	\label{alg: optimization}
    \setlength{\tabcolsep}{1.0mm}
	\begin{algorithmic}[1]
		\REQUIRE The consecutive tasks $\mathcal{T} = \{\mathcal{T}^t\}_{t=1}^T$, $\{\alpha_1, \alpha_2\}$; 
		\FOR {$t=1, 2, \cdots, T$}
		\WHILE{not converged}
		\IF {$t=1$}
		\STATE Obtain a mini-batch $\{\mf{x}_i^t, \mf{y}_i^t\}_{i=1}^b\in\mathcal{T}^t$; 
		\STATE Update $\Theta^t$, $\mathbf{E}_0^t$ via optimizing $\mathcal{L}_{\mr{CE}}$ in Eq.~\eqref{eq: classification_loss};
		\ELSE
		\STATE Obtain a mini-batch $\{\mf{x}_i^t, \mf{y}_i^t\}_{i=1}^b\in\mathcal{T}^t\cup\mathcal{M}$; 
		\STATE Update $\Theta^t$, $\mathbf{E}_0^t$ via optimizing $\mathcal{L}_{\mr{obj}}$ in Eq.~\eqref{eq: overall_objective};
		\ENDIF
		\ENDWHILE
		\STATE  Update memory $\mathcal{M}$ via following \cite{Rebuffi_2017_CVPR};
		\STATE  Store $\Theta^t$ as $\Theta^{t-1}$, and $\mathbf{E}_0^t$ for next initialization.
		\ENDFOR \\
	\end{algorithmic}
\end{algorithm}

\subsection{Qualitative Analysis of Convergence}\label{sec: convergence}
\begin{figure}[ht]
	\centering
	\includegraphics[trim = 48mm 52mm 49mm 52mm, clip, width=236pt, height=118pt]
	{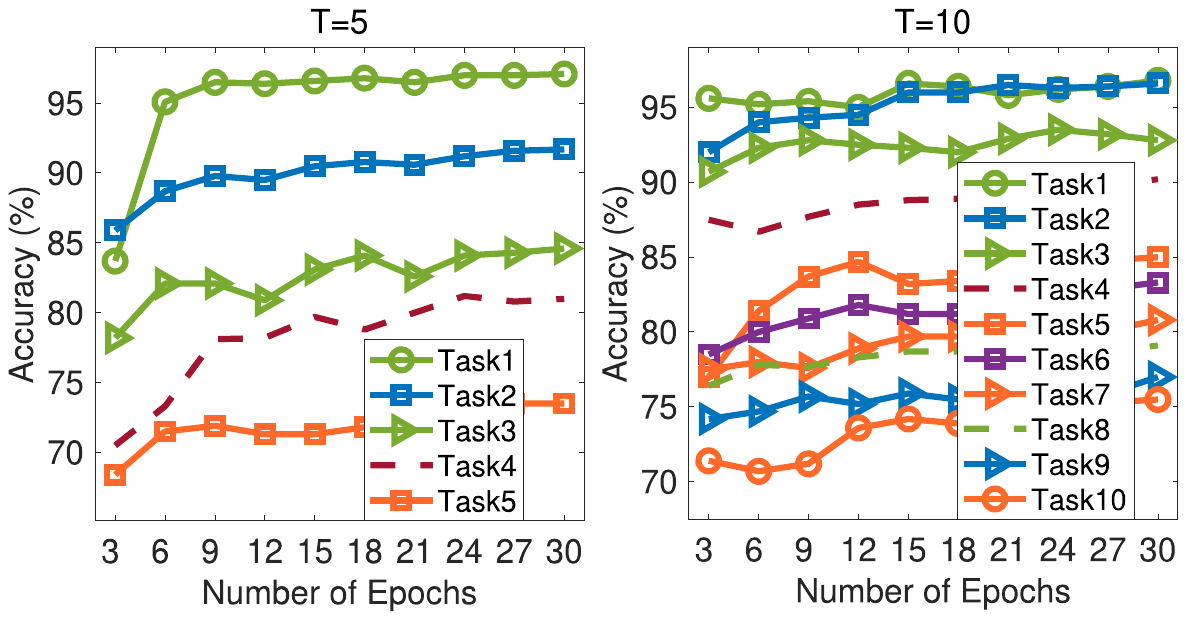}
	\vspace{-15pt}
	\caption{Convergence curves on ImageNet-100 \cite{5206848}, when we set $\mathcal{M}=2000, \mathcal{B}=0\%$ for $T=\{5, 10\}$. }
	\label{fig: conver_ImageNet100}
\end{figure}

As presented in Fig.~\ref{fig: conver_ImageNet100}, we introduce some qualitative convergence results in terms of top-1 accuracy  for each incremental task on ImageNet-100 \cite{5206848}. From these curves, we can observe that the accuracy in each increment task increases gradually until convergence as the training process. It shows that our proposed HFC model has robust convergence performance for each incremental task. More importantly, the convergence speed is fast via optimizing the proposed HFC model with only few training epochs. It also illustrates the effectiveness of our proposed model to address heterogeneous catastrophic forgetting on old classes.

\begin{table*}[t]
\centering
\setlength{\tabcolsep}{1.2mm}
\caption{Performance of each incremental task on ImageNet-100 \cite{5206848} in terms of top-1 accuracy, when $\mathcal{M}=2000$, $T\!=\!10$ and $\mathcal{B}\!=\!0\%$.  }
\scalebox{0.87}{
\begin{tabular}{l|cc|cccccccccc|cc}
	\toprule
\makecell[c]{\multirow{2}{*}{Comparison Methods}} & \multirow{2}{*}{Backbone} & \multirow{2}{*}{\#Params} & \multicolumn{12}{c}{$\mathcal{B}=0\%$}  \\
	  &  &  & 10 & 20 & 30 & 40 & 50 & 60 & 70 & 80 & 90 & 100 & Avg. & Imp. \\
	\midrule
 iCaRL \cite{Rebuffi_2017_CVPR} (CVPR'2017) & ViT-Base & 85.10M  &96.4 &95.4 &91.5 &86.4 &81.3 &78.6 &76.6 &74.8 &72.8 &70.4 &82.4 &$\Uparrow$3.0  \\
	BiC \cite{wu2019large} (CVPR'2019) & ViT-Base & 85.10M &\textcolor{deepred}{\textbf{97.8}} &96.1 &89.5 &86.8 &79.0 &76.8 &73.5 &72.7 &67.3 &62.9 &80.2 &$\Uparrow$5.2 \\
	PODNet \cite{10.1007/978-3-030-58565-5_6} (ECCV'2020) & ViT-Base & 85.10M &\textcolor{deepred}{\textbf{97.8}} &96.1 &89.5 &86.9 &79.1 &76.9 &73.6 &71.9 &66.4 &63.1 &80.1 &$\Uparrow$5.3  \\
	SS-IL \cite{Ahn_2021_ICCV} (ICCV'2021) & ViT-Base& 85.10M &96.6 &96.2 &91.1 &86.9 &81.2 &77.7 &76.1 &75.0 &68.9 &67.0 &81.7 &$\Uparrow$3.7   \\
 PODNet \cite{10.1007/978-3-030-58565-5_6} + CSCCT \cite{10.1007/978-3-031-19812-0_7} (ECCV'2022) & ViT-Base & 85.10M &96.8 &94.0 &90.3 &85.6 &79.0 &75.6 &75.1 &74.8 &70.4 &68.1 &81.0 &$\Uparrow$4.4 \\
 FOSTER \cite{10.1007/978-3-031-19806-9_23} (ECCV'2022) & ViT-Base & 85.10M &96.4 &96.4 &92.0 &88.0 &83.5 &81.1 &79.1 &77.2 &74.8 &73.0 &84.0 & $\Uparrow$1.4 \\
        AFC \cite{Kang_2022_CVPR} (CVPR'2022) &ViT-Base & 85.10M &96.8 &96.7 &89.5 &86.3 &79.7 &77.5 &75.5 &74.7 &70.0 &65.6 &81.2 &$\Uparrow$4.2  \\
	DyTox \cite{Douillard_2022_CVPR} (CVPR'2022) & ViT-Base & 85.10M &96.4 &95.7 &92.2 &88.2 &83.0 &79.5 &77.0 &75.3 &70.4 &66.8 &83.4 &$\Uparrow$2.0  \\
        \midrule
        \rowcolor{lightgray}
        \textbf{HFC} (\textbf{Ours}) & ViT-Base & 85.10M   &96.8 &\textcolor{deepred}{\textbf{96.6}} &\textcolor{deepred}{\textbf{92.6}} &\textcolor{deepred}{\textbf{89.2}} &\textcolor{deepred}{\textbf{85.1}} &\textcolor{deepred}{\textbf{82.7}} &\textcolor{deepred}{\textbf{80.5}} &\textcolor{deepred}{\textbf{79.0}} &\textcolor{deepred}{\textbf{76.8}} &\textcolor{deepred}{\textbf{75.5}} &\textcolor{deepred}{\textbf{85.4}}  &\textbf{$\mathrm{-}$}   \\
        \midrule
        Upper Bound & ViT-Base & 85.10M &\textbf{$\mathrm{-}$} &\textbf{$\mathrm{-}$} &\textbf{$\mathrm{-}$} &\textbf{$\mathrm{-}$} &\textbf{$\mathrm{-}$} &\textbf{$\mathrm{-}$} &\textbf{$\mathrm{-}$} &\textbf{$\mathrm{-}$} &\textbf{$\mathrm{-}$} &\textbf{$\mathrm{-}$}&86.3 &\textbf{$\mathrm{-}$}\\
 \bottomrule
 
\end{tabular}}
\label{tab: comparison_exp_img100}
\vspace{-5pt}
\end{table*}

\begin{table*}[t]
\centering
\setlength{\tabcolsep}{1.2mm}
\caption{Performance of each incremental task on ImageNet-1000 \cite{5206848} in terms of top-1 accuracy, when $\mathcal{M}=2000$, $T\!=\!10$ and $\mathcal{B}\!=\!0\%$. }
\scalebox{0.87}{
\begin{tabular}{l|cc|cccccccccc|cc}
	\toprule
\makecell[c]{\multirow{2}{*}{Comparison Methods}} & \multirow{2}{*}{Backbone} & \multirow{2}{*}{\#Params} & \multicolumn{12}{c}{$\mathcal{B}=0\%$}  \\
	  &  &  & 10 & 20 & 30 & 40 & 50 & 60 & 70 & 80 & 90 & 100 & Avg. & Imp. \\
	\midrule
 iCaRL \cite{Rebuffi_2017_CVPR} (CVPR'2017) & ViT-Base & 85.10M  &90.7 &83.5 &78.2 &75.2 &72.4 &70.0 &67.1 &64.3 &61.8 &61.0 &72.4 &$\Uparrow$4.0   \\
	BiC \cite{wu2019large} (CVPR'2019) & ViT-Base & 85.10M &90.7 &84.4 &76.4 &71.9 &66.6 &61.1 &57.4 &53.5 &50.8 &47.4 &66.0 &$\Uparrow$10.4 \\
	PODNet \cite{10.1007/978-3-030-58565-5_6} (ECCV'2020) & ViT-Base & 85.10M &84.4 &79.1 &75.8 &72.7 &71.4 &69.0 &67.1 &65.8 &64.7 &63.5 &71.3 &$\Uparrow$5.1   \\
	SS-IL \cite{Ahn_2021_ICCV} (ICCV'2021) & ViT-Base& 85.10M &85.5 &85.3 &78.4 &78.0 &76.0 &73.2 &71.3 &69.8 &66.5 &63.7 &74.8 &$\Uparrow$1.6  \\
 PODNet \cite{10.1007/978-3-030-58565-5_6} + CSCCT \cite{10.1007/978-3-031-19812-0_7} (ECCV'2022) & ViT-Base & 85.10M &90.3 &67.6 &62.7 &62.2 &60.9 &60.0 &58.8 &57.7 &56.4 &55.7 &63.2 &$\Uparrow$13.2 \\
 FOSTER \cite{10.1007/978-3-031-19806-9_23} (ECCV'2022) & ViT-Base & 85.10M  &90.9 &84.8 &79.7 &77.0 &74.2 &72.1 &69.4 &67.3 &64.5 &63.4 &74.3 &$\Uparrow$2.1 \\
        AFC \cite{Kang_2022_CVPR} (CVPR'2022) &ViT-Base & 85.10M &90.5 &86.3 &80.1 &76.3 &72.2 &67.8 &64.6 &61.6 &58.3 &56.0 &71.4 &$\Uparrow$5.0 \\
	DyTox \cite{Douillard_2022_CVPR} (CVPR'2022) & ViT-Base & 85.10M &\textcolor{deepred}{\textbf{91.5}} &\textcolor{deepred}{\textbf{88.1}} &\textcolor{deepred}{\textbf{83.1}} &\textcolor{deepred}{\textbf{79.8}} &76.3 &72.2 &68.6 &65.7 &62.4 &59.4 &74.7 &$\Uparrow$1.7  \\
        \midrule
        \rowcolor{lightgray}
        \textbf{HFC} (\textbf{Ours}) & ViT-Base & 85.10M  &90.7 &85.9 &81.6 &79.3 &\textcolor{deepred}{\textbf{76.6}} &\textcolor{deepred}{\textbf{74.4}} &\textcolor{deepred}{\textbf{71.7}} &\textcolor{deepred}{\textbf{69.9}} &\textcolor{deepred}{\textbf{67.9}} &\textcolor{deepred}{\textbf{66.0}} &\textcolor{deepred}{\textbf{76.4}} &\textbf{$\mathrm{-}$}   \\
        \midrule
        Upper Bound & ViT-Base & 85.10M &\textbf{$\mathrm{-}$} &\textbf{$\mathrm{-}$} &\textbf{$\mathrm{-}$} &\textbf{$\mathrm{-}$} &\textbf{$\mathrm{-}$} &\textbf{$\mathrm{-}$} &\textbf{$\mathrm{-}$} &\textbf{$\mathrm{-}$} &\textbf{$\mathrm{-}$} &\textbf{$\mathrm{-}$}&86.3 &\textbf{$\mathrm{-}$}\\
 \bottomrule
 
\end{tabular}}
\label{tab: comparison_exp_img1000}
\vspace{-10pt}
\end{table*}

\begin{table*}[t]
\centering
\setlength{\tabcolsep}{1.80mm}
\caption{Performance on CIFAR-100 \cite{krizhevsky2009learning} ($T=10$),  when we apply \textbf{Ours$^\ddag$} into existing distillation-based CIL methods and set $\mathcal{M}=2000, \mathcal{B}=0\%$. \textbf{Ours$^\ddag$} denotes the proposed plug-and-play losses $\mathcal{L}_{\rm{FC}}$ and $\mathcal{L}_{\rm{RD}}$. }
\scalebox{0.81}{
\begin{tabular}{l|cc|cccccccccc|cc}
	\toprule
\makecell[c]{\multirow{2}{*}{Comparison Methods}} & \multirow{2}{*}{Backbone} & \multirow{2}{*}{\#Params} & \multicolumn{12}{c}{$\mathcal{B}=0\%$}   \\
	  &  &  & 10 & 20 & 30 & 40 & 50 & 60  & 70 & 80 & 90 & 100 & Avg. & Imp. \\
	\midrule
iCaRL \cite{Rebuffi_2017_CVPR} (CVPR'2017) & ResNet-32 & 0.46M &84.2 &77.3 &73.0 &68.4 &63.5 &60.5 &58.3 &54.5 &52.3 &47.4 &63.9 &$\Uparrow$1.8  \\
        \rowcolor{lightgray}
        iCaRL \cite{Rebuffi_2017_CVPR} + \textbf{Ours}$^\ddag$ & ResNet-32 & 0.46M &\textcolor{deepred}{\textbf{84.2}}  &\textcolor{deepred}{\textbf{78.6}}  &\textcolor{deepred}{\textbf{74.9}}  &\textcolor{deepred}{\textbf{69.8}}  &\textcolor{deepred}{\textbf{65.6}}  &\textcolor{deepred}{\textbf{62.5}}  &\textcolor{deepred}{\textbf{60.3}} &\textcolor{deepred}{\textbf{55.4}}  &\textcolor{deepred}{\textbf{54.2}}  &\textcolor{deepred}{\textbf{51.4}}  &\textcolor{deepred}{\textbf{65.7}} &\textbf{$\mathrm{-}$} \\
        \midrule
	BiC \cite{wu2019large} (CVPR'2019) & ResNet-32 & 0.46M &88.9 &70.1 &56.0 &45.7 &43.9 &46.6 &41.6 &39.9 &39.1 &35.8 &50.8 &$\Uparrow$3.0   \\
  \rowcolor{lightgray} 
        BiC \cite{wu2019large} + \textbf{Ours}$^\ddag$ & ResNet-32 & 0.46M &\textcolor{deepred}{\textbf{88.9}}  &\textcolor{deepred}{\textbf{72.1}}  &\textcolor{deepred}{\textbf{58.1}}  &\textcolor{deepred}{\textbf{49.0}}  &\textcolor{deepred}{\textbf{47.7}}  &\textcolor{deepred}{\textbf{49.4}}  &\textcolor{deepred}{\textbf{44.8}}  &\textcolor{deepred}{\textbf{44.0}}  &\textcolor{deepred}{\textbf{43.5}}  &\textcolor{deepred}{\textbf{40.5}} &\textcolor{deepred}{\textbf{53.8}} &\textbf{$\mathrm{-}$}   \\
        \midrule
	PODNet \cite{10.1007/978-3-030-58565-5_6} (ECCV'2020) & ResNet-32 & 0.46M &85.7 &73.0 &63.8 &56.4 &53.7 &48.3 &45.2 &42.0 &39.0 &38.1 &54.5 &$\Uparrow$5.5    \\
 \rowcolor{lightgray}
        PODNet \cite{10.1007/978-3-030-58565-5_6} + \textbf{Ours}$^\ddag$ & ResNet-32 & 0.46M &\textcolor{deepred}{\textbf{89.5}} &\textcolor{deepred}{\textbf{77.8}} &\textcolor{deepred}{\textbf{69.6}} &\textcolor{deepred}{\textbf{62.6}} &\textcolor{deepred}{\textbf{57.2}} &\textcolor{deepred}{\textbf{55.6}} &\textcolor{deepred}{\textbf{51.1}} &\textcolor{deepred}{\textbf{49.6}} &\textcolor{deepred}{\textbf{45.9}} &\textcolor{deepred}{\textbf{41.2}} &\textcolor{deepred}{\textbf{60.0}}   &\textbf{$\mathrm{-}$}   \\
        \midrule
	SS-IL \cite{Ahn_2021_ICCV} (ICCV'2021) & ResNet-32 & 0.46M &84.7 &67.3 &64.5 &60.2 &57.4 &54.5 &53.5 &50.9 &49.8 &47.6 &59.0 &$\Uparrow$5.9    \\
 \rowcolor{lightgray}
 SS-IL \cite{Ahn_2021_ICCV} + \textbf{Ours}$^\ddag$ & ResNet-32 & 0.46M &\textcolor{deepred}{\textbf{86.0}} &\textcolor{deepred}{\textbf{78.2}} &\textcolor{deepred}{\textbf{74.2}} &\textcolor{deepred}{\textbf{68.9}} &\textcolor{deepred}{\textbf{64.8}} &\textcolor{deepred}{\textbf{61.7}} &\textcolor{deepred}{\textbf{59.6}} &\textcolor{deepred}{\textbf{54.3}} &\textcolor{deepred}{\textbf{52.9}} &\textcolor{deepred}{\textbf{48.0}} &\textcolor{deepred}{\textbf{64.9}} &\textbf{$\mathrm{-}$}   \\
 \midrule
  PODNet \cite{10.1007/978-3-030-58565-5_6} + CSCCT \cite{10.1007/978-3-031-19812-0_7} (ECCV'2022) & ResNet-32 & 0.46M  &85.7 &73.0 &63.7 &56.3 &53.6 &48.1 &44.9 &41.7 &38.7 &37.8 &54.3 &$\Uparrow$4.2 \\
   \rowcolor{lightgray}
   PODNet \cite{10.1007/978-3-030-58565-5_6} + CSCCT \cite{10.1007/978-3-031-19812-0_7} + \textbf{Ours}$^\ddag$ & ResNet-32 & 0.46M &\textcolor{deepred}{\textbf{85.7}} &\textcolor{deepred}{\textbf{75.0}} &\textcolor{deepred}{\textbf{67.7}}&\textcolor{deepred}{\textbf{60.9}} &\textcolor{deepred}{\textbf{57.9}} &\textcolor{deepred}{\textbf{53.2}} &\textcolor{deepred}{\textbf{50.4}} &\textcolor{deepred}{\textbf{46.9}} &\textcolor{deepred}{\textbf{44.0}} &\textcolor{deepred}{\textbf{43.5}} &\textcolor{deepred}{\textbf{58.5}}   &\textbf{$\mathrm{-}$}  \\
   \midrule
 FOSTER \cite{10.1007/978-3-031-19806-9_23} (ECCV'2022) & ResNet-32 & 0.46M &91.9 &82.0 &75.7 &71.0 &67.8 &65.4 &61.8 &59.5 &58.4 &53.7 &68.7 &$\Uparrow$1.2  \\
 \rowcolor{lightgray}
   FOSTER \cite{10.1007/978-3-031-19806-9_23} + \textbf{Ours}$^\ddag$ & ResNet-32 & 0.46M &\textcolor{deepred}{\textbf{91.9}} &\textcolor{deepred}{\textbf{82.2}} &\textcolor{deepred}{\textbf{77.2}} &\textcolor{deepred}{\textbf{71.5}} &\textcolor{deepred}{\textbf{68.4}} &\textcolor{deepred}{\textbf{67.4}} &\textcolor{deepred}{\textbf{63.3}} &\textcolor{deepred}{\textbf{61.6}}&\textcolor{deepred}{\textbf{59.5}} &\textcolor{deepred}{\textbf{56.0}} &\textcolor{deepred}{\textbf{69.9}}  &\textbf{$\mathrm{-}$}   \\
        \midrule
        AFC \cite{Kang_2022_CVPR} (CVPR'2022) & ResNet-32 & 0.46M &\textcolor{deepred}{\textbf{90.9}} &76.5 &65.7 &57.4 &52.8 &51.8 &47.4 &45.5 &42.8 &40.2 &57.1 &$\Uparrow$6.7   \\
        \rowcolor{lightgray}
   AFC \cite{Kang_2022_CVPR} + \textbf{Ours}$^\ddag$ & ResNet-32 & 0.46M  &88.7 &\textcolor{deepred}{\textbf{81.2}} &\textcolor{deepred}{\textbf{72.3}} &\textcolor{deepred}{\textbf{65.9}} &\textcolor{deepred}{\textbf{62.6}} &\textcolor{deepred}{\textbf{60.2}} &\textcolor{deepred}{\textbf{56.2}} &\textcolor{deepred}{\textbf{53.4}} &\textcolor{deepred}{\textbf{50.2}} &\textcolor{deepred}{\textbf{46.8}} &\textcolor{deepred}{\textbf{63.8}} &\textbf{$\mathrm{-}$}  \\
   \midrule
   Upper Bound & ResNet-32 & 0.46M &\textbf{$\mathrm{-}$} &\textbf{$\mathrm{-}$} &\textbf{$\mathrm{-}$} &\textbf{$\mathrm{-}$}  &\textbf{$\mathrm{-}$} &\textbf{$\mathrm{-}$} &\textbf{$\mathrm{-}$} &\textbf{$\mathrm{-}$} &\textbf{$\mathrm{-}$} &\textbf{$\mathrm{-}$} &76.6 &\textbf{$\mathrm{-}$}\\
	\bottomrule
    \bottomrule
 
DyTox \cite{Douillard_2022_CVPR} (CVPR'2022) & ViT-Tiny & 10.71M &\textcolor{deepred}{\textbf{91.9}} &85.0 &80.0 &74.8 &73.4 &71.4 &68.2 &65.9 &63.9 &61.0 &73.5 &$\Uparrow$0.5  \\ 
\rowcolor{lightgray}
DyTox \cite{Douillard_2022_CVPR} + \textbf{Ours}$^\ddag$ & ViT-Tiny & 10.71M&91.7 &\textcolor{deepred}{\textbf{86.0}} &\textcolor{deepred}{\textbf{80.5}} &\textcolor{deepred}{\textbf{74.8}} &\textcolor{deepred}{\textbf{73.7}} &\textcolor{deepred}{\textbf{72.0}} &\textcolor{deepred}{\textbf{68.5}} &\textcolor{deepred}{\textbf{66.7}} &\textcolor{deepred}{\textbf{64.7}} &\textcolor{deepred}{\textbf{61.1}} &\textcolor{deepred}{\textbf{74.0}} &\textbf{$\mathrm{-}$}  \\
\midrule
Upper Bound & ViT-Tiny & 10.71M &\textbf{$\mathrm{-}$} &\textbf{$\mathrm{-}$} &\textbf{$\mathrm{-}$} &\textbf{$\mathrm{-}$}  &\textbf{$\mathrm{-}$} &\textbf{$\mathrm{-}$} &\textbf{$\mathrm{-}$} &\textbf{$\mathrm{-}$} &\textbf{$\mathrm{-}$} &\textbf{$\mathrm{-}$} &76.1 &\textbf{$\mathrm{-}$}\\
 \bottomrule
 
\end{tabular}}
\label{tab: two_losses_cifar}
\vspace{-10pt}
\end{table*}

\end{document}